\documentclass{article}
\usepackage[nonatbib]{neurips_2024}

\usepackage[utf8]{inputenc}
\usepackage[T1]{fontenc}
\usepackage{hyperref}
\usepackage{url}
\usepackage{booktabs}
\usepackage{amsfonts}
\usepackage{nicefrac}
\usepackage{microtype}
\usepackage{comment}
\usepackage{xcolor}
\usepackage{longtable}
\usepackage{listings}
\usepackage{listings}
\usepackage{xcolor}
\usepackage[pdftex]{graphicx}
\usepackage{amsmath}
\usepackage{booktabs}
\usepackage{float}
\usepackage{placeins}
\usepackage{amsmath} % Include this in the preamble
\usepackage{tabularx}      % For automatic column width adjustment
\title{EQUATOR: A Deterministic Framework for Evaluating LLM Reasoning with Open-Ended Questions.   \# v1.0.0-beta }

\author{%
Raymond Bernard\thanks{
about the author (raymondbernard.github.io)} \\
Independent Researcher \\
Kitchener, On N2M534 \\
\texttt{ray.bernard@outlook.com} \\
\And
Shaina Raza PhD \\
Vector Institute \\
Toronto, ON M5G 1M1  \\
\texttt{shaina.raza@vectorinstitute.ai} \\
\And
Subhabrata Das, PhD \\
JP Morgan Chase VP. \\
New York, New York \\
\texttt{sd2957@columbia.edu} \\
\And
Rahul  Murugan \\
Columbia University \\
New York, New York, NY 10027 \\
\texttt{rmm2292@columbia.edu } \\
}

\begin{document}

\maketitle

\begin{abstract}
Despite the remarkable coherence of Large Language Models (LLMs), existing evaluation methods often suffer from fluency bias and rely heavily on multiple-choice formats, making it difficult to assess factual accuracy and complex reasoning effectively. LLMs thus frequently generate factually inaccurate responses, especially in complex reasoning tasks, highlighting two prominent challenges: (1) the inadequacy of existing methods to evaluate reasoning and factual accuracy effectively, and (2) the reliance on human evaluators for nuanced judgment, as illustrated by Williams and Huckle (2024)\cite{sean_williams_c6f85baf}, who found manual grading indispensable despite automated grading advancements.

To address evaluation gaps in open-ended reasoning tasks, we introduce the EQUATOR Evaluator (Evaluation of Question Answering Thoroughness in Open-ended Reasoning). This framework combines deterministic scoring with a focus on factual accuracy and robust reasoning assessment. Using a vector database, EQUATOR pairs open-ended questions with human-evaluated answers, enabling more precise and scalable evaluations. In practice, EQUATOR significantly reduces reliance on human evaluators for scoring and improves scalability compared to Williams and Huckle's (2004)\cite{sean_williams_c6f85baf} methods.

Our results demonstrate that this framework significantly outperforms traditional multiple-choice evaluations while maintaining high accuracy standards. Additionally, we introduce an automated evaluation process leveraging smaller, locally hosted LLMs.   We used  LLaMA 3.2B, running on the  \href{https://ollama.com/}{Ollama} binaries to streamline our assessments. This work establishes a new paradigm for evaluating LLM performance, emphasizing factual accuracy and reasoning ability, and provides a robust methodological foundation for future research.

\end{abstract}
\newpage
\section{Introduction}
Can a deterministic scoring framework, specifically the EQUATOR Evaluator, provide a more objective, reliable, and scalable method for evaluating the factual accuracy and reasoning capabilities of LLMs in response to open-ended questions compared to traditional evaluation methods?
The paper challenges the reliance on traditional evaluation methods, such as multiple-choice formats, which are often susceptible to biases like fluency bias and lack a thorough assessment of reasoning abilities.
The paper explores the hypothesis that a deterministic scoring framework, utilizing a vector database of human-evaluated answers, can mitigate these limitations and offer a more precise evaluation of LLM performance, particularly in complex reasoning tasks where factual accuracy is paramount.

As Large Language Models (LLMs) become increasingly integrated into various applications, evaluating their responses accurately is critical. Traditional evaluation methods often fail to assess model performance, mainly because LLMs generate fluent and logically coherent responses that are stochastic and hallucinate, which means they are usually factually incorrect \cite{xinghua_zhang_e4ab2763}. Human and automated evaluators also assign higher scores based on the fluency of reactions, the position of the response, and frequency rather than their factual accuracy.

The challenges in evaluating LLM responses and biases have motivated researchers to develop more deterministic approaches. Some key factors that have driven this need include likelihood bias  (overrate sentences with higher likelihoods while underrating those with lower likelihoods)\cite{ohi2024likelihood},  position bias (prioritize content based on its position within the given context) \cite{wang2024eliminating}, lack of reference text \cite{sheng2024reference}, limitations in semantic understanding, and the need for explain-ability \cite{wu2023style}. LLMs can also exhibit various types of biases \cite{raza2024developing}, including demographic, cultural, linguistic, and temporal biases, which arise from factors such as training data, algorithmic choices, and model architecture.

Williams and Huckle (2024) \cite{sean_williams_c6f85baf} highlighted that LLMs frequently fail at open-ended reasoning and logic questions, even when these questions are straightforward for humans. This issue underscores the necessity for a more rigorous evaluation system that prioritizes factual accuracy over persuasive ability. 
Our work addresses this gap by introducing a deterministic scoring framework emphasizing factual correctness in LLM outputs. \cite{cheng_han_chiang_2023}. 

% \textcolor{red}{Raymond, how did this approach address, please write here, the below line is methodology }
We leverage a vector database constructed from questions and corresponding human-evaluated answers; we provide a reliable source of truth that focuses on the logical and factual content of responses.

Our work differentiates itself from previous works, such as \cite{zhihui_xie_2024,katuka2024investigating} by introducing a deterministic scoring framework that leverages a vector database. This framework addresses the limitations of traditional methods, which often rely on multiple-choice formats or human-evaluated answers, by providing a scalable and flexible source of truth. Unlike prior approaches, we incorporate smaller large language models (LLMs) as automated evaluators, combining computational efficiency with advanced language understanding to emphasize factual accuracy over linguistic fluency.
While prior research has explored semantic similarity and embeddings for answer evaluation, these methods are often influenced by the fluency and persuasiveness of responses, potentially introducing bias. By contrast, our approach mitigates such biases by prioritizing objective correctness, ensuring reliable and scalable evaluations.

% Real-world consequences of LLM hallucinations and reasoning failures, especially in high-stakes applications. *** 
The real-world consequences of hallucinations and reasoning failures in large language models (LLMs) are profound, particularly in high-stakes applications such as healthcare, finance, legal systems, and governance. LLM hallucinations—when models generate false or fabricated information—can lead to decisions based on incorrect or misleading data, with potentially catastrophic outcomes. In healthcare, for instance, an LLM providing inaccurate advice on medical treatments or diagnostics could jeopardize patient safety. Similarly, in the financial sector, reasoning failures in tasks such as risk assessment, portfolio management, or fraud detection can result in significant monetary losses, regulatory violations, or even systemic instability. The stakes are equally high in legal systems, where flawed reasoning could misguide legal professionals or decision-makers, undermining justice and fairness.

A deterministic framework like EQUATOR is important because of its ability to systematically evaluate LLM reasoning under open-ended and context-dependent scenarios, where the risk of errors is amplified. By identifying the limitations of LLMs in reasoning and hallucination-prone areas, EQUATOR helps mitigate the risks associated with deploying these models in real-world, high-stakes environments. For instance, the framework's structured evaluation ensures that models meet rigorous benchmarks before applying in sensitive domains. This not only improves the reliability of AI systems but also fosters trust among stakeholders who rely on them for critical decision-making. EQUATOR’s deterministic and reproducible approach to evaluation is therefore a vital step in bridging the gap between LLM capabilities and the demands of real-world applications, ensuring that their deployment aligns with safety, accuracy, and accountability standards.

A key challenge would be designing practical and scalable techniques. For example, how can we develop differentially private model training approaches 
that are applicable for billions or trillions of parameters in generative AI models? How can we ensure privacy of end users when leveraging inputs from end users as part of retraining of LLMs (using, say, PATE-like approaches)? Considering the importance of
high quality datasets for evaluating LLMs for truthfulness, bias, robustness, safety, and related dimensions, and the challenges with obtaining such datasets in highly sensitive domains such as healthcare, how do we develop practical and feasible approaches for differentially private synthetic data generation, potentially leveraging a combination of sensitive datasets (e.g., patient health records and clinical notes) and publicly available datasets along with the ability to generate data by querying powerful LLMs?

% of efforts that often rely on multiple-choice formats or human-evaluated answers, providing a scalable and flexible source of truth. Unlike previous methods, we integrate smaller LLMs as automated evaluators, combining efficiency with advanced language understanding to focus on factual accuracy over linguistic fluency.
% While prior research has utilized semantic similarity and embeddings for answer evaluation, these methods can be biased by the fluency and persuasiveness of responses.

The specific contributions of this work are:
% \textbf{Our approach differs in several novel contributions{}: 
 
 \begin{enumerate}
     \item We employ a deterministic scoring system using a vector database to enhance scalability and flexibility for evaluating diverse, open-ended responses.
     \item We achieve efficient and consistent assessments by integrating smaller LLMs as automated evaluators. This approach reduces computational demands and mitigates biases by focusing on deterministic scoring with open-ended questions and a vector database of human-evaluated answers.
     \item Our method explicitly addresses the challenges LLMs face in reasoning and logic tasks, emphasizing factual accuracy without using multiple choices as an option.
 \end{enumerate}

Our goal is to assess the performance of state-of-the-art models on complex reasoning tasks by comparing their capabilities to human performance. This approach provides a nuanced understanding of model strengths and weaknesses in areas where they have historically underperformed, helping us identify when and if these models reach human-level problem-solving abilities.

The empirical analysis compared benchmarking results from two studies using deterministic scoring frameworks and multiple-choice question formats evaluated by humans and LLMs. Findings revealed that deterministic scoring produced significantly lower mean scores than traditional methods, with a large effect size (\(d = 1.07\)) in the 2024-09-13 dataset and an even larger effect size (\(d = 2.85\)) in the 2024-06-12 dataset, confirming statistically significant differences (\(p < 0.005\) and \(p < 0.0001\), respectively). These results highlight the advantages of deterministic frameworks in reducing ambiguity and ensuring consistent evaluation, showcasing their robustness for logical reasoning benchmarks.
\section{Related Work}
The EQUATOR Evaluator builds on and advances research in evaluating large language models (LLMs), particularly in areas where existing benchmarks fall short. A notable precursor is the work by Williams and Huckle (2024) \cite{sean_williams_c6f85baf}, which highlighted frequent failures of LLMs in handling open-ended reasoning and logic questions—even those considered straightforward for humans. This underscores the need for rigorous evaluation systems that prioritize factual accuracy over fluency and persuasive ability.

\subsection{Limitations of Existing Benchmarks}

Many traditional evaluation frameworks, such as SuperGLUE (Wang et al., 2019)\cite{wang2019superglue} and BIG-bench (Srivastava et al., 2022)\cite{srivastava2022beyond}, focus on tasks that LLMs are already proficient in, like linguistic fluency and simple reasoning. While these benchmarks have advanced the field, they rely heavily on multiple-choice scenarios, which present inherent limitations:

Simplistic Question Formats: Multiple-choice questions often reduce the complexity of open-ended reasoning to a set of predefined answers, limiting the scope of model evaluation.

Bias Toward Fluent Responses: LLMs frequently score well on fluency-focused benchmarks, even when their answers are factually incorrect.

Performance Saturation: Models rapidly approach human-level scores on multiple-choice benchmarks, making it challenging to distinguish between truly innovative architectures and incremental improvements.

\subsection{Innovations in EQUATOR}

EQUATOR sets itself apart by addressing these limitations through a deterministic scoring framework designed for open-ended questions. This framework introduces several innovations:

Open-Ended Evaluation: Unlike prior benchmarks, EQUATOR eliminates the reliance on multiple-choice formats. By evaluating free-form responses, it captures the nuanced reasoning capabilities and factual accuracy of LLMs.

Deterministic Scoring: The framework employs a vector database of human-evaluated answers, ensuring objective and reproducible assessments. This contrasts with traditional methods that rely on subjective human grading or automated metrics susceptible to fluency bias.

Integration of Smaller LLMs: EQUATOR leverages smaller, locally hosted models like LLaMA 3.2B as automated evaluators. This approach balances computational efficiency with advanced language understanding, reducing the overhead associated with large-scale models.

\subsection{Addressing Hallucinations and Reasoning Failures}

The prevalence of hallucinations—where LLMs generate fluent but factually incorrect responses—is a significant challenge in model evaluation. Prior studies, such as those by Zhang et al. (2023)\cite{zhang2024multiple} and Ohi et al. (2024)\cite{ohi2024likelihood}, have proposed methods to mitigate evaluation bias, but these often fall short in open-ended scenarios. EQUATOR’s deterministic framework directly tackles this issue by:

Pairing questions with human-evaluated answers stored in a vector database.

Using cosine similarity to objectively evaluate semantic closeness between model-generated responses and the reference answers.

Adopting binary scoring criteria, which prioritize factual correctness over reasoning or fluency.

\subsection{Comparative Analysis}

In contrast to prior works like SuperGLUE \cite{wang2019superglue} and BIG-bench \cite{srivastava2022beyond}, which emphasize linguistic and logical fluency, EQUATOR focuses on factual accuracy and robust reasoning. While previous research has explored embeddings and semantic similarity for evaluation, these methods often favor persuasive but inaccurate responses. EQUATOR mitigates such biases by prioritizing correctness over fluency. Employing smaller LLMs to automate evaluation, ensuring scalability and cost-efficiency.
Avoiding the pitfalls of traditional human grading by standardizing deterministic criteria.

\subsection{Real-World Implications}

EQUATOR’s emphasis on open-ended QA evaluation has significant implications for high-stakes applications like medicine, law, and education. Providing a scalable, unbiased, and efficient evaluation framework paves the way for more reliable LLM development, ultimately contributing to the creation of models that align more closely with human reasoning and factual accuracy.

In summary, EQUATOR represents a paradigm shift in LLM evaluation, moving beyond the constraints of multiple-choice benchmarks to offer a more comprehensive and reliable assessment framework. This work establishes a strong foundation for future research aimed at addressing the limitations of existing methodologies and advancing the capabilities of LLMs in real-world scenarios.
\newpage

\begin{figure}
    \centering
    \includegraphics[width=1\linewidth]{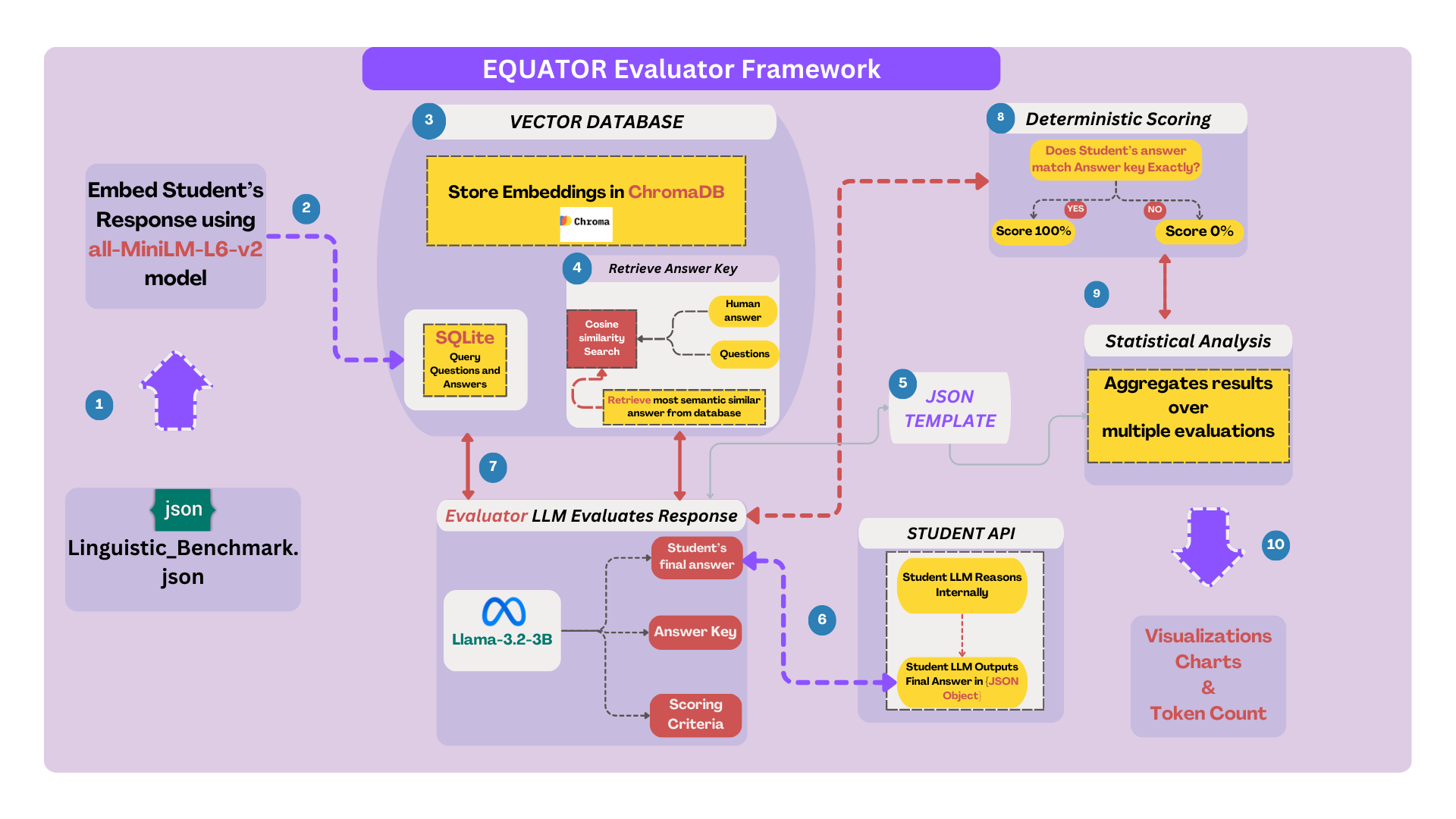}
    \caption{EQUATOR Evaluator Framework - Drawn by Rahul Murugan}
    \label{fig: EQUATOR Evaluator Framework}
\end{figure}

\section{EQUATOR Evaluation Framework}
Our evaluation framework is shown in Figure \ref{fig: EQUATOR Evaluator Framework}, which consists of the below steps. 

\begin{enumerate}
    \item  - Questions and Human Answers are pulled from the linguistic benchmark.json file.
    \begin{enumerate}
        \item  - The linguistic benchmark.json is embedded using the all-MiniLM-L6-v2 from Ollama
    \end{enumerate}
    \item - Vector Database is created using ChromaDB, which also creates an SQLite instance.
    \item  - The SQLite db is used to iterate the questions in the Vector DB to feed the LLM Evaluator.
    \item  -  Perform a vector document search using cosine similarity to retrieve the closest semantic human answer
    \item Create JSON Template and rounds directory structure.
    \item  - The question is sent to an API along with instructions to the "student" to return the final answer in a JSON Object, i.e. {'student answer':'my answer'} We are careful to give time and space for the student to reason through the answer.
    \item  - The EQUATOR Evaluator will score the student based on the scoring criteria and its answer key from the Vector database.
    \item  -  Each time we pass the JSON file with the score, the student's answer, and the EQUATOR Evaluator's rational
\\  * Note: for each question: go back to steps 5 through 8 until we have completed the specified number of questions and runs.
    \item  -  We then aggregate the results over multiple evaluations to produce several statistical views from the data.
    \item - We produce the visual charts and tables and calculate the tokens used for the total evaluation.
\end{enumerate}

\newpage

\subsection{Preliminaries} To formalize the deterministic scoring framework, we define key terms used throughout this work. These definitions are summarized in Table~\ref{tab:preliminaries}.
\begin{table}[h!]
\small
\centering
\caption{Preliminaries and Definitions}
\label{tab:preliminaries}
\begin{tabular}{|p{3cm}|p{10cm}|}
\hline
\textbf{Term} & \textbf{Definition} \\ \hline
\textbf{Question (\( Q_s \))} & The open-ended question presented to the student or the LLM for response generation. \\ \hline
\textbf{Student Answer (\( A_s \))} & The response provided by the student or the LLM being evaluated. \\ \hline
\textbf{Answer Keys (\( K \))} & A set of human-evaluated correct answers used as references during the scoring process. \\ \hline
\textbf{Embedding} & A dense, high-dimensional vector representation of textual inputs that captures semantic similarity. \\ \hline
\textbf{Cosine Similarity} & A metric that measures the cosine of the angle between two embeddings, indicating their semantic closeness. \\ \hline
\textbf{Vector Database} & A specialized database designed to store and retrieve embeddings efficiently for similarity searches. \\ \hline
\end{tabular}
\end{table}

\subsection{Vector Database Construction}

We construct a vector database using embeddings of open-ended questions and their corresponding human-evaluated correct answers. This database serves as the source of truth for evaluating LLM responses.
For embedding both questions and human-evaluated answers, we utilized the \verb|ll-minilm| model from the Ollama library \cite{reimers2019sentencebert}. This model generates 384-dimensional sentence embeddings, balancing efficiency and performance in capturing semantic similarity. 
Text inputs were preprocessed to maintain consistency by serializing them into JSON format. Embeddings were generated using a local embedding service implemented with Ollama and stored in a ChromaDB vector database. Cosine similarity was employed to measure the semantic closeness between embeddings during retrieval and evaluation.

\subsection{Components}
The evaluation framework relies on the following components:
\begin{enumerate}
    \item \textbf{Semantic Similarity Search:} Vector databases compare student responses to reference answers by embedding them as vectors and performing similarity searches to determine their proximity to high-quality responses. This is analogous to how systems like VectorRAG retrieve relevant information for LLMs.

    \item \textbf{Flexibility and Scalability:} Vector databases are more straightforward to scale and manage than graph databases, making them suitable for handling large datasets and frequently updated evaluation criteria.

    \item \textbf{Integration with LLMs:} Vector databases integrate with LLMs for tasks such as retrieval, ranking, and clustering, enabling nuanced analysis \cite{sarmah2024hybridrag} and accurate evaluation of student responses.

    \item \textbf{Response Evaluation:} LLM-generated responses are embedded using the same methodology and compared against the vector database using cosine similarity. A high similarity score indicates that the response closely matches the correct answer.

    \item \textbf{Threshold Determination:} Instead of setting an empirical similarity threshold, we retrieve the most semantically similar answer from the vector database as the answer key. The evaluator LLM compares the student's response directly with this key. According to predefined criteria, a score of 100\% is assigned if the answers match exactly, and 0\% otherwise. This binary evaluation eliminates the need for manual threshold tuning while leveraging the evaluator LLM's capabilities.
\end{enumerate}

\subsection{Deterministic Scoring Framework}

Given a question \( Q_s \), a student's answer \( A_s \), and a set of answer keys \( K = \{K_1, K_2, \dots, K_n\} \), the deterministic scoring framework is defined through the following steps:

\textbf{Step 1: Embed Inputs} Each input is transformed into its corresponding embedding representation:
\begin{align*}
Q_e & = \text{Embed}(Q_s) \quad \text{(Embedding of the Question)} \\
A_e & = \text{Embed}(A_s) \quad \text{(Embedding of the Student's Answer)} \\
K_e & = \{\text{Embed}(K_i)\ |\ K_i \in K\} \quad \text{(Embeddings of the Answer Keys)}
\end{align*}

\textbf{Step 2: Retrieve Closest Answer Key}
The closest answer key \( K_{\text{best}} \) is identified by maximizing the cosine similarity between the embedding of the student's answer \( A_e \) and the embeddings of the answer keys \( K_e \):
\[
K_{\text{best}} = \arg\max_{K_i \in K} \text{Sim}(A_e, K_i)
\]
Where the cosine similarity function is defined as:
\[
\text{Sim}(x, y) = \frac{x \cdot y}{\|x\| \|y\|}
\]

\textbf{Step 3: Evaluate Match Based on Instructions}
The match between the student's answer and the best-matching answer key is evaluated based on a predefined similarity threshold \( \tau \):
\[
\text{Match}(A_e, K_{\text{best}}) = 
\begin{cases} 
1, & \text{if } \text{Sim}(A_e, K_{\text{best}}) \geq \tau \\
0, & \text{otherwise}
\end{cases}
\]

\textbf{Step 4: Assign Score} The final score \( S \) is computed as:
\[
S = \text{Match}(A_e, K_{\text{best}}) \times 100\%
\]

\textbf{Output} The framework outputs the final score \( S \), which represents the deterministic evaluation of the student's response against the provided set of answer keys.

% \newpage

\subsection{LLM in an Evaluator Role}
To automate the evaluation process and eliminate human bias, we employed a smaller LLM, specifically LLaMA 3.2B, as the evaluator. The evaluator LLM is prompted to act as a strict high school teacher who grades based solely on the provided answer key, ignoring reasoning or persuasive elements in the student's response. 

The scoring criteria are binary:
\begin{itemize}
    \item \textbf{100\%}: Assigned when the student's final answer exactly matches the answer key.
    \item \textbf{0\%}: Assigned when the student's final answer does not match the answer key. No partial credit is allowed.
\end{itemize}
The evaluator LLM returns the score and a brief explanation in a structured JSON format. This method ensures a consistent and objective evaluation process while leveraging the efficiency of smaller LLMs to achieve scalability.

\subsubsection{Student and Evaluator Prompts}

The "student LLM" (LLM under evaluation) is provided with the following prompt:

\begin{lstlisting}
{
  "role": "system",
  "content": "You are a student who is being tested. Please follow the directions given exactly. You are welcome to reason through the question. You must return only your final answer in a JSON Object, for example: {'student_answer': '<My final answer here>'}"
}
\end{lstlisting}

The Evaluator System Prompt assumes the persona of a strict high school teacher who has advanced to a Ph.D.-level professor. This persona grades the student strictly based on the answer key, overlooking logical reasoning or persuasive arguments that deviate from the provided correct answers.

Evaluator system configuration:
\begin{lstlisting}

  evaluator_system_prompt = [
            {   "role": "system",
                "content": (
"""You are a strict high school teacher 
your role is to grade strictly based on the provided answer key provided. Only examine the final answer given by the student in a JSON Object {'student_answer:<My final Answer>} The student's answer = Answer key given!! No partial credit allowed! Do not consider the logical arguments which are not relevant to the question"""
                    + system_prompt
                ),
            },
            
\end{lstlisting}

% \newpage 

\subsubsection{Evaluator System Prompt with Scoring Criteria}

{\small}
\begin{lstlisting}
 system_prompt = """
    SCORING CRITERIA
    100%: The response must match the answer key given, even if the student used bad reasoning and logic to arrive at the final answer. 
    0%: The answer key does NOT match the one given. No partial credit allowed!

    TASK
    Evaluate whether the STUDENT Answer matches the answer key given. If it does, assign a score of 100%. Otherwise, you must assign a score of 0%.  Provide a very short explanation of why.  
    Just focus on the student's final answer!  Give full credit to the student if the final answer matches the answer key. Don't overthink this. Also do not evaluate based on the quality, logical reasoning, even if it is very persuasive!
    Only consider the answer key as the source of truth.  Your job is at risk if you do not follow our instructions.  If the Answer Key matches the student's answer, you must assign a score of 0%; no partial credit is allowed.    
    Return a JSON object explaining why the student got his score.  Keep it do less than two sentences {"evaluation": "<explanation>"}
    Return a JSON object with the score as a percentage. Example:
    {"score": 0}"""

\end{lstlisting}
To replace human evaluators and enhance scalability, we employ smaller, locally hosted LLMs like LLaMA 3.2B to perform the embedding and comparison tasks. These models are efficient and reduce computational overhead.
\subsection{Taxonomy}

Our study utilizes a dataset of open-ended questions structured according to the taxonomy proposed by Williams and Huckle (2024) \cite{sean_williams_c6f85baf}.

The dataset includes various question types that challenge LLM reasoning capabilities:

\begin{itemize}
    \item \textbf{Puzzle}: Logic puzzles requiring basic reasoning.
    \item \textbf{Spatial}: Questions about the arrangement or positions of objects.
    \item \textbf{Relational}: Inferences about relationships or hierarchies.
    \item \textbf{Counting}: Simple numerical calculations.
    \item \textbf{Linguistic}: Language understanding with specific constraints.
    \item \textbf{Popular Science}: Questions testing common scientific misconceptions.
\end{itemize}

The Williams and Huckle Toxonmony \cite{sean_williams_c6f85baf} does provide answers derived from human input. \ref{appendix:Evaluation Table}Our goal is to expand the question and answer set to achieve statistical significance. Currently, we have increased our collection to approximately 1,005 questions, including those that require lateral thinking. We will be publishing an update on our new website featuring the EQUATOR Framework.

% \subsection{Experiments and Results }
% \subsection{Experimental Setting}
% \textcolor{red}{please write experimental setting heres}

% \textcolor{blue}(TODO: 
% goal of this evaluation is...
% 1- Ray can you please add hardware, memory , software in 1 line.
% 2- which baseline models choose, just name them here.
% 3- evaluation question (what are you trying to do, asking their response?? or what, just 1 line .
% 4. metrics names.
% 5. benchmark datasets you are usng, just name all in 1 line.
% )

\section{Experiments and Results for two studies}

To evaluate the effectiveness of our deterministic scoring framework, we conducted 2 experiments using datasets from Williams and Huckle (2024) \cite{sean_williams_c6f85baf}. Specifically, we utilized their\href{https://github.com/autogenai/easy-problems-that-llms-get-wrong/tree/main/2024-09-13-Multi-Benchmark/auto_eval_outputs}{2024-09-13-Multi-Benchmark dataset} for multiple-choice questions and the \href{https://github.com/autogenai/easy-problems-that-llms-get-wrong/tree/main/2024-06-12-Benchmark/auto_eval_outputs}{2024-06-12-Benchmark dataset} for open-ended question answering (QA). This allowed us to compare the performance of various state-of-the-art LLMs under traditional evaluation methods and our proposed framework.

\section{Experiment 1 - 2024-9-13-Multiple-Choice-Benchmark}

% \subsection{First Experiment}
In the first experiment,  the EQUATOR framework was applied to the multiple-choice answers given by the LLMS on the Williams and Huckle 9-13 Multiple-choice benchmark.   The other results are directly from the Williams and Huckle results for the 9-13 multiple choice benchmark \cite{sean_williams_c6f85baf}

\begin{table}[ht]
\centering
% \begin{adjustbox}
\begin{tabular}{lrrrrrr}
\toprule
\textbf{Model} & \textbf{Mean} & \textbf{Std Dev} & \textbf{Z Err} & \textbf{CI Lower} & \textbf{CI Upper} & \textbf{Count} \\
\midrule
Human-Eval & 86.0 & 0.0 & 0.0 & 93.0 & 78.0 & 0 \\
gpt-4o & 57.47 & 34.10 & 3.86 & 53.67 & 61.27 & 300.0 \\
gemini-1\_5-pro & 53.93 & 35.30 & 3.99 & 49.87 & 57.87 & 300.0 \\
claude-3-opus-20240229 & 52.6 & 35.95 & 4.07 & 48.67 & 56.60 & 300.0 \\
gpt-4-turbo-preview & 50.13 & 34.72 & 3.93 & 46.27 & 54.07 & 300.0 \\
mistral-large-latest & 49.33 & 34.60 & 3.91 & 45.47 & 53.20 & 300.0 \\
meta\_llama3-70b-instr-v1\_0 & 46.8 & 32.45 & 3.67 & 43.2 & 50.53 & 300.0 \\
open-mixtral-8x22b & 44.2 & 33.44 & 3.78 & 40.47 & 48.07 & 300.0 \\
gpt-4o-*E& 41.10 & 49.29 & 5.58 & 35.49 & 46.74 & 300.0 \\
gemini-1\_0-pro & 39.67 & 34.83 & 3.94 & 35.8 & 43.6 & 300.0 \\
claude-3-opus-20240229-*E & 38.26 & 48.68 & 5.51 & 32.55 & 43.81 & 300.0 \\
mistral-large-latest-*E& 29.77 & 45.80 & 5.18 & 24.67 & 34.90 & 300.0 \\
gemini-1\_0-pro-*E& 27.70 & 44.83 & 5.07 & 22.71 & 32.88 & 300.0 \\
command-r & 27.07 & 32.48 & 3.68 & 23.47 & 30.73 & 300.0 \\
gpt-4-turbo-preview-*E& 23.96 & 42.76 & 4.84 & 19.18 & 28.92 & 300.0 \\
meta\_llama3-70b-inst-v1-*E& 23.01 & 42.06 & 4.76 & 18.18 & 27.93 & 300.0 \\
open-mixtral-8x22b-*E& 20.67 & 40.52 & 4.59 & 16.08 & 25.54 & 300.0 \\
gemini-1\_5-pro-*E& 19.93 & 40.02 & 4.53 & 15.44 & 24.66 & 300.0 \\
command-r-*E& 13.38 & 34.10 & 3.86 & 9.67 & 17.33 & 300.0 \\
\bottomrule
\end{tabular}
% \end{adjustbox}
\caption{Williams and Huckle 9-13 Multiple Choice QA scored with the EQUATOR framework.\\ *E = EQUATOR}
\label{tab:model_scores1}
\end{table}

The results in Table \ref{tab:model_scores1} offers a comparative analysis of various LLMs evaluated on the Williams and Huckle 9-13 Multiple-Choice benchmark using the EQUATOR framework. 
% The table includes metrics such as mean score, standard deviation, Z-inter error, confidence interval lower and upper bounds, and the number of evaluations for each model.
This analysis establishes a baseline by applying the EQUATOR framework to multiple-choice answers generated by different LLMs with the goal to reveal performance variations across various models.

Our expectation is that the EQUATOR framework will provide more stringent and accurate scores compared to the scoring criteria developed by Williams and Huckle \ref{appendix: scoring criteria}, even though their scoring was performed by humans. Williams and Huckle made efforts to automate the scoring process but ultimately found it to be too inaccurate. They noted that multiple-choice questions are the most deterministic and reliable for evaluation since there is a clear set answer to measure against. However, open-ended questions can more effectively reveal illogical and inconsistent behavior, though they are challenging to evaluate.

For open-ended questions (non multiple-choice), Williams and Huckle \href{https://github.com/autogenai/easy-problems-that-llms-get-wrong?tab=readme-ov-file#most-accurate-results}{GITHUB}
recommend that a person manually mark the LLM responses instead of relying on the scores auto-generated in the auto\_eval\_outputs folder (which are by default marked by GPT\-4o). Users can edit the scores in the auto\_eva\_outputs JSON files directly and then re\-run the "generate\_statistics" execution step in the main.ipynb notebook to obtain the final results. This manual adjustment process, as employed by the authors for their paper, resulted in significantly lower scores compared to the less reliable LLM-based auto evaluation.

We have observed that when LLMs are used as automatic evaluators, they tend to assign significantly higher scores, often providing partial credit based on their reasoning processes. Additionally, these models struggle to reason through complex logic and rely heavily on their pretraining rather than genuine analytical capabilities. The EQUATOR framework aims to address these limitations by offering a more robust and accurate evaluation method.  For enhanced visualization, we have provided several graphical views of our first experiment in the appendix \ref{appendix: graphical views 9-13}

% \subsection{Experiment \# 1 - 9-13 Multiple Choice w Human Evaluation Benchmarks}
\begin{comment}
What is the purpose of this section, i commented, undo it if i missed.

\subsection{Multiple Choice w Human Evaluation Benchmarks}

For comparison, we also evaluated the models using traditional multiple-choice questions and scoring methods. This allowed us to assess the performance differences between multiple-choice and open-ended question evaluations.

We use the data sets runs used by Williams and Huckle(2024)  \cite{sean_williams_c6f85baf}. Specifically, the 9-13-Multi-benchmark evaluates multiple-choice questions.  
Click here  \href{https://github.com/autogenai/easy-problems-that-llms-get-wrong/tree/main/2024-09-13-Multi-Benchmark/auto_eval_outputs}{easy-problems-that-llms-get-wrong} 
\end{comment}
\newpage 

\subsection{Statistical}

To validate the significance of our results using the real data provided by Williams and Huckle (2024) \cite{sean_williams_c6f85baf}, We conducted detailed statistical analyses comparing the STOA models' performance under traditional evaluation methods as conducted by Williams and Huckle (2024) \cite{sean_williams_c6f85baf} vs our New Equator deterministic scoring framework. Our objective was to determine whether the differences observed were statistically significant and to quantify the magnitude of these differences.

\paragraph{Paired t-tests} 
We performed paired t-tests for each model to compare the mean scores obtained under the two evaluation methods. The paired t-test is appropriate here because we compare two related sets of scores---the same models evaluated on the same questions but scored using two different frameworks.
%  \paragraph{Data Preparation}

We paired each model's mean scores under the traditional evaluation and the deterministic scoring framework. The data is summarized in Table~\ref{tab:paired_means1}.

\begin{table}[ht]
\centering
\caption{Paired Mean Scores for Each Model}
\label{tab:paired_means1}
\begin{tabular}{lcc}
\toprule
\textbf{Model} & \textbf{Mean Score} & \textbf{EQUATOR Deterministic Mean Score} \\
\midrule
GPT-4o & 42.00 & 40.61 \\
GPT-4-turbo-preview & 36.67 & 26.39 \\
Claude-3-opus-20240229 & 44.33 & 37.00 \\
O1-preview & 72.00 & 35.23 \\
Meta-Llama-3-1-70B-Instruct & 34.33 & 23.95 \\
Meta-Llama-3-1-405B-Instruct & 54.67 & 33.78 \\
GPT-4o-mini-2024-07-18 & 33.33 & 30.50 \\
Gemini-1.5-pro & 31.00 & 22.41 \\
Mistral-large-latest & 29.67 & 26.17 \\
Meta\_llama3-70b-instruct-v1.0 & 29.00 & 22.21 \\
Open-mixtral-8x22b & 27.00 & 21.72 \\
Claude-3-5-sonnet-20240620 & 50.67 & 20.75 \\
Command-r-plus & 18.67 & 11.33 \\
\bottomrule
\end{tabular}
\end{table}

\paragraph{Calculating Differences}

For each model, we calculated the difference in mean scores:

\[
\text{Difference} = \text{Traditional Mean Score} - \text{Deterministic Mean Score}
\]

The differences are shown in Table~\ref{tab:mean_differences1}.

\begin{table}[ht]
\centering
\caption{Differences in Mean Scores}
\label{tab:mean_differences1}
\begin{tabular}{lc}
\toprule
\textbf{Model} & \textbf{Difference} \\
\midrule
GPT-4o & 1.39 \\
GPT-4-turbo-preview & 10.28 \\
Claude-3-opus-20240229 & 7.33 \\
O1-preview & 36.77 \\
Meta-Llama-3-1-70B-Instruct & 10.38 \\
Meta-Llama-3-1-405B-Instruct & 20.88 \\
GPT-4o-mini-2024-07-18 & 2.83 \\
Gemini-1.5-pro & 8.59 \\
Mistral-large-latest & 3.49 \\
Meta\_llama3-70b-instruct-v1.0 & 6.79 \\
Open-mixtral-8x22b & 5.28 \\
Claude-3-5-sonnet-20240620 & 29.92 \\
Command-r-plus & 7.33 \\
\bottomrule
\end{tabular}
\end{table}

For a detailed account of the math, please refer to the appendix. \ref{appendix:stat9-13}

\subsubsection{Effect Size (Cohen's d )}

To quantify the magnitude of the observed difference, we calculated Cohen's \( d \):

\[
d = \frac{\bar{d}}{s_d} = \frac{11.59}{10.86} \approx 1.07
\]

\paragraph{Interpretation} \ref{cohen:cohensd9-13}

\paragraph{Normality Assumption}

With \( n = 13 \)T, the sample size is relatively small. However, the Central Limit Theorem assures that the sampling distribution of the mean difference is approximately normal.

\paragraph{Independence}

We assume that the differences between the traditional and deterministic scores for each model are independent of those for other models.

\subsubsection{Discussion for Multiple-Choice human scoring vs EQUATOR Evaluator scoring} 
\label{cohen:cohensd9-13}

The statistical analyses confirm that our deterministic scoring framework leads to significantly lower mean scores than traditional evaluation methods. The significant t-test results (\( p < 0.005 \)) indicate that these differences are unlikely due to random chance.

The large effect size (Cohen's \( d \approx 1.07 \)) highlights the substantial impact of the scoring framework on model performance assessments. This supports our assertion that traditional evaluation methods may overestimate LLM performance by not adequately penalizing factual inaccuracies.

\begin{figure}
\centering
\subsubsection{\textbf{Experiment 1- Explanation of the Cohen’s d Effect Sizes Chart} } \ref{appendix:Chart Analysis}
\includegraphics[width=1\linewidth]{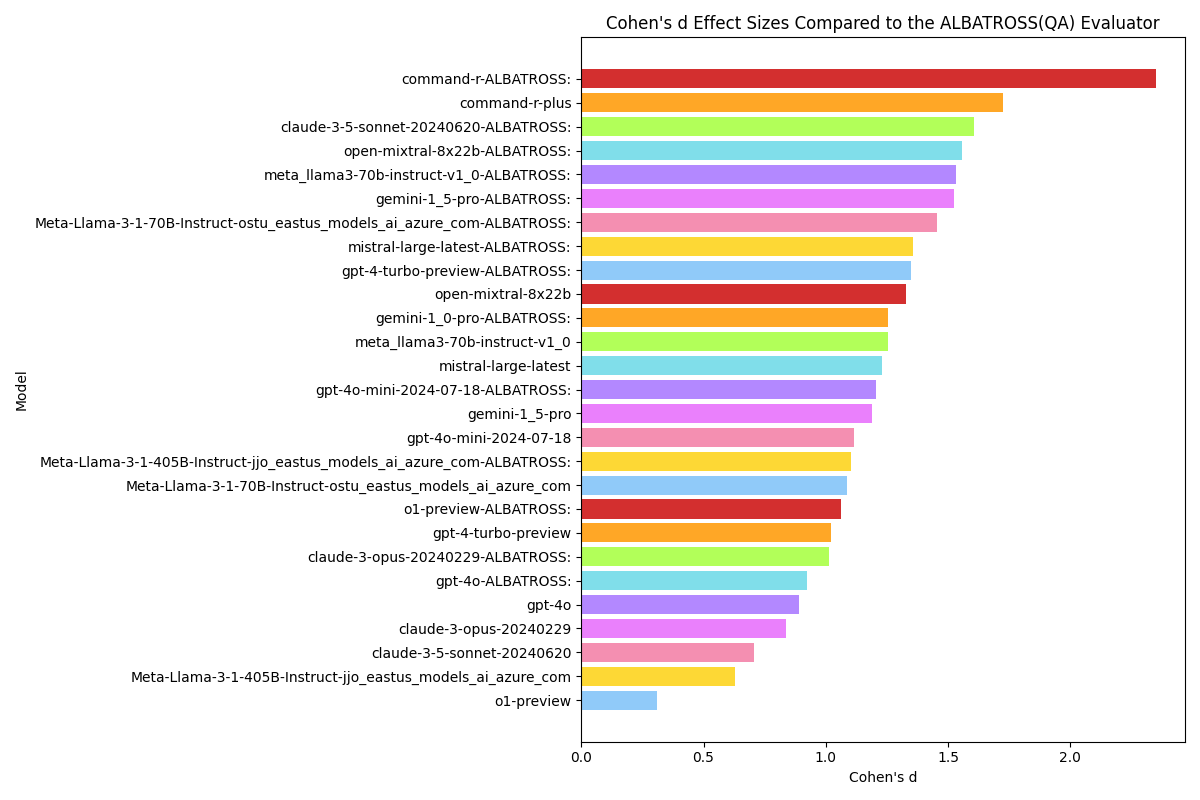}
\caption{Cohen’s d Effect Sizes Chart}
\label{fig: Cohen’s d Effect Sizes Chart}
\end{figure}

\FloatBarrier
\newpage

\section{Experiment - 2:  2014-6-12-Benchmarks *Open Ended Questions}

\begin{table}[b!]
    \centering
    \small % Reduce font size; options include \footnotesize, \scriptsize
    \cite{sean_williams_c6f85baf} \\
    Williams and Huckle 6-12 Open-ended QA w/ Human score QA with the EQUATOR Evaluator
    \vspace{0.5em} % Add some vertical space between the description and the table
    % \begin{adjustbox}
        \begin{tabular}{lrrrrrr}
            \toprule
            \textbf{Model} & \textbf{Mean} & \textbf{Std Dev} & \textbf{Z Inter Error} & \textbf{CI Lower} & \textbf{CI Upper} & \textbf{Count} \\
            \midrule
            Human-Eval & 86.0 & 0.0 & 0.0 & 93.0 & 78.0 & 0 \\
            gpt-4o & 57.47 & 34.10 & 3.86 & 53.67 & 61.27 & 300.0 \\
            gemini-1\_5-pro & 53.93 & 35.30 & 3.99 & 49.87 & 57.87 & 300.0 \\
            claude-3-opus-20240229 & 52.6 & 35.95 & 4.07 & 48.67 & 56.60 & 300.0 \\
            gpt-4-turbo-preview & 50.13 & 34.72 & 3.93 & 46.27 & 54.07 & 300.0 \\
            mistral-large-latest & 49.33 & 34.60 & 3.91 & 45.47 & 53.20 & 300.0 \\
            meta\_llama3-70b-instr-v1\_0 & 46.8 & 32.45 & 3.67 & 43.2 & 50.53 & 300.0 \\
            open-mixtral-8x22b & 44.2 & 33.44 & 3.78 & 40.47 & 48.07 & 300.0 \\
            gpt-4o-Falcon & 41.10 & 49.29 & 5.58 & 35.49 & 46.74 & 300.0 \\
            gemini-1\_0-pro & 39.67 & 34.83 & 3.94 & 35.8 & 43.6 & 300.0 \\
            claude-3-opus-20240229-*E & 38.26 & 48.68 & 5.51 & 32.55 & 43.81 & 300.0 \\
            mistral-large-latest-*E & 29.77 & 45.80 & 5.18 & 24.67 & 34.90 & 300.0 \\
            gemini-1\_0-pro-*E & 27.70 & 44.83 & 5.07 & 22.71 & 32.88 & 300.0 \\
            command-r & 27.07 & 32.48 & 3.68 & 23.47 & 30.73 & 300.0 \\
            gpt-4-turbo-preview-*E & 23.96 & 42.76 & 4.84 & 19.18 & 28.92 & 300.0 \\
            meta\_llama3-70b-instr-v1-*E & 23.01 & 42.06 & 4.76 & 18.18 & 27.93 & 300.0 \\
            open-mixtral-8x22b-*E & 20.67 & 40.52 & 4.59 & 16.08 & 25.54 & 300.0 \\
            gemini-1\_5-pro-*E & 19.93 & 40.02 & 4.53 & 15.44 & 24.66 & 300.0 \\
            command-r-*E & 13.38 & 34.10 & 3.86 & 9.67 & 17.33 & 300.0 \\
            \bottomrule
        \end{tabular}
    % \end{adjustbox}
    \caption{Model Comparison Scores (*E = EQUATOR)}
    \label{tab:model_scores}
\end{table}

Open-ended questions are defined by their free-form nature, absence of multiple-choice options, and the requirement for the large language model (LLM) to generate original, unrestricted responses. These responses are uniquely evaluated using the EQUATOR framework, ensuring a comprehensive assessment of factual accuracy and reasoning capabilities.

To validate the significance of our results using we used the open-ended questions and answers provided by the Williams and Huckle study.\href{https://github.com/autogenai/easy-problems-that-llms-get-wrong/tree/main/2024-06-10-Benchmark}{2024-06-12 Benchmark} 

\begin{figure}
\centering
\section{\textbf{Experiment -2 Explanation of the Cohen’s d Effect Sizes Chart} } \ref{appendix:Chart Analysis}
\includegraphics[width=1\linewidth]{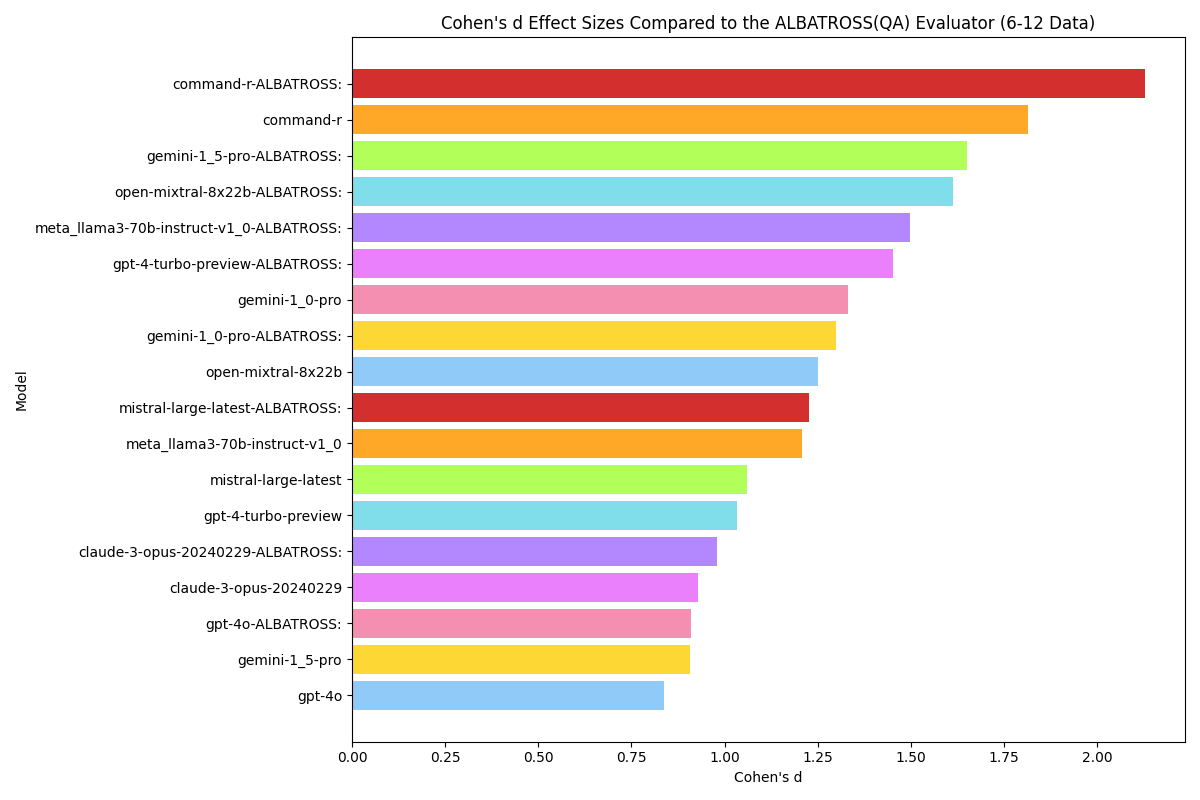}
\caption{Cohen’s d Effect Sizes Chart}
\label{fig: Cohen’s d Effect Sizes 2014-6-12-Benchmark}
\end{figure}

We conducted detailed statistical analyses comparing the models' performance under traditional evaluation methods and our deterministic scoring EQUATOR framework. Our objective was to determine whether the differences observed were statistically significant and to quantify the magnitude of these differences; we used the Cohen D \ref{fig: Cohen’s d Effect Sizes 2014-6-12-Benchmark}

\textbf{Paired t-tests}

We performed paired t-tests for each model to compare the mean scores obtained under the two evaluation methods. The paired t-test is appropriate here because we are comparing two related sets of scores—the same models evaluated on the same questions but scored using two different frameworks.

\paragraph{Data Preparation}

We paired each model's mean scores under the traditional evaluation and the deterministic scoring framework. The data are summarized in Table~\ref{tab:paired_means}.

\begin{table}[b!]
\centering
\caption{Paired Mean Scores for Each Model}
\label{tab:paired_means}
\begin{tabular}{lcc}
\toprule
\textbf{Model} & \textbf{Mean Score} & \textbf{EQUATOR Deterministic Mean Score} \\
\midrule
GPT-4o & 57.47 & 41.10 \\
Gemini-1\_5-pro & 53.93 & 19.93 \\
Claude-3-opus-20240229 & 52.60 & 38.26 \\
GPT-4-turbo-preview & 50.13 & 23.96 \\
Mistral-large-latest & 49.33 & 29.77 \\
Meta\_llama3-70b-instruct-v1\_0 & 46.80 & 23.01 \\
Open-mixtral-8x22b & 44.20 & 20.67 \\
Gemini-1\_0-pro & 39.67 & 27.70 \\
Command-r & 27.07 & 13.38 \\
\bottomrule
\end{tabular}
\end{table}

\paragraph{Calculating Differences}

For each model, we calculated the difference in mean scores:

\[
\text{Difference} = \text{Traditional Mean Score} - \text{Deterministic Mean Score}
\]

The differences are shown in Table~\ref{tab:mean_differences}.

\begin{table}[ht]
\centering
\caption{Differences in Mean Scores}
\label{tab:mean_differences}
\begin{tabular}{lc}
\toprule
\textbf{Model} & \textbf{Difference} \\
\midrule
GPT-4o & $16.37$ \\
Gemini-1\_5-pro & $34.00$ \\
Claude-3-opus-20240229 & $14.34$ \\
GPT-4-turbo-preview & $26.18$ \\
Mistral-large-latest & $19.57$ \\
Meta\_llama3-70b-instruct-v1\_0 & $23.79$ \\
Open-mixtral-8x22b & $23.53$ \\
Gemini-1\_0-pro & $11.96$ \\
Command-r & $13.69$ \\
\bottomrule
\end{tabular}
\end{table}

For a detailed view of the maths, please refer to our appendix \ref{appendix:Stats-6-12}

\section{Effect Size (Cohen's d}
\label{effect-size-6-12}
To quantify the magnitude of the observed difference, we calculated Cohen's \( d \):

\[
d = \frac{\bar{d}}{s_d} = \frac{20.38}{7.16} \approx 2.85
\]

\paragraph{Interpretation}

According to Cohen's conventions:

\begin{itemize}
    \item Small effect: \( d = 0.2 \)
    \item Medium effect: \( d = 0.5 \)
    \item Large effect: \( d = 0.8 \)
\end{itemize}

A Cohen's \( d \) of approximately \( 2.85 \) indicates a very large effect size, suggesting that the deterministic scoring framework has a substantial impact on the evaluation outcomes.

\subsection{Assumptions and Considerations}

\paragraph{Normality Assumption}

With \( n = 9 \), the sample size is modest. However, the Central Limit Theorem assures that the sampling distribution of the mean difference is approximately normal.

\paragraph{Independence}

We assume that the differences between the traditional and deterministic scores for each model are independent of those for other models.

\subsubsection{Final Comments Experiment-2  2024-6-12-Benchmark}

The statistical analyses confirm that our deterministic scoring framework leads to significantly lower mean scores than traditional evaluation methods. The significant t-test results (\( p < 0.0001 \)) indicate that these differences are highly unlikely due to random chance.

The very large effect size (Cohen's \( d \approx 2.85 \)) highlights the substantial impact of the scoring framework on model performance assessments. This supports our assertion that traditional evaluation methods may overestimate LLM performance by not adequately penalizing factual inaccuracies.

\section{Empirical analysis for both experiments:}

We analyzed the differences in benchmarking results from the studies conducted by Williams and Huckles using the \href{https://github.com/autogenai/easy-problems-that-llms-get-wrong/tree/main/2024-09-13-Multi-Benchmark/auto_eval_outputs}{2024-09-13-Multi-Benchmark dataset}.

The authors converted open-ended questions into multiple-choice format and used human evaluators to score responses based on criteria detailed in Appendix~\ref{appendix: scoring criteria}. They also experimented with an auto-scoring script powered by an LLM as the evaluator. Their findings indicated that the most accurate results were achieved when questions were formatted as multiple-choice and scored by humans, as this approach minimized ambiguity and ensured deterministic scoring. Additional details are available \href{https://github.com/autogenai/easy-problems-that-llms-get-wrong/tree/main?tab=readme-ov-file#most-accurate-results}{here}.

This benchmark represents the latest in multiple-choice QA evaluation, including assessments of OpenAI’s o1 reasoning model. Our evaluator was able to grade 30 logical and reasoning questions accurately, with results visually confirmed and documented in Appendix~\ref{appendix:Evaluation Table}.

Using the same statistical methodology as Williams and Huckles (2024) \cite{sean_williams_c6f85baf}, we evaluated all ten runs across both benchmarks. Statistical analysis revealed that our deterministic scoring framework produced significantly lower mean scores than those reported in the Williams and Huckles study ($p \approx 0.0026$). A t-test confirmed that this difference was statistically significant ($p < 0.005$), suggesting it was unlikely to have occurred by chance.

To quantify the effect size of the observed differences, we calculated Cohen’s $d$, which was 1.07—a large effect size ($d \geq 0.8$).

We then applied our LLM Evaluator to re-score the \href{https://github.com/autogenai/easy-problems-that-llms-get-wrong/tree/main/2024-06-10-Benchmark}{2024-06-12 Benchmark} originally scored by Williams and Huckles. The results showed a statistically significant difference between traditional and deterministic scoring frameworks ($p < 0.0001$), allowing us to reject the null hypothesis that the mean scores under the two methods were the same. Cohen’s $d$ for this second evaluation was 2.85, indicating an even larger effect size.

\section*{Comparative Table of Both  Studies}

\begin{table}[h!]
    \centering
    \resizebox{\linewidth}{!}{%
        \begin{tabular}{|l|l|l|}
            \hline
            \textbf{Aspect} & \textbf{2024-09-13-Multi-Benchmark} & \textbf{2024-06-12-Benchmark} \\ \hline
            \textbf{Evaluation Method} & Deterministic scoring framework & Deterministic scoring framework \\ \hline
            \textbf{Scoring Approach} & LLM Evaluator and human scoring & LLM Evaluator \\ \hline
            \textbf{Statistical Significance} & $p \approx 0.0026$ & $p < 0.0001$ \\ \hline
            \textbf{Effect Size (Cohen’s $d$)} & 1.07 (large) & 2.85 (very large) \\ \hline
            \textbf{Main Finding} & Lower mean scores with deterministic framework & Larger differences in mean scores confirmed \\ \hline
        \end{tabular}
    }
    \caption{Comparison of Benchmark Studies}
    \label{tab:comparison}
\end{table}

% \subsubsection{Implications}

% \begin{itemize}
%     \item \textbf{Evaluation Practices:} The findings suggest a need to re-evaluate current practices in assessing LLMs to ensure that evaluations accurately reflect models' abilities to produce factually correct responses.
%     \item \textbf{Model Development:} Developers should prioritize improving models' factual accuracy and reasoning capabilities over fluency or persuasive language.
%     \item \textbf{Future Research:} Further studies could expand the sample size of models and explore different tasks to generalize these findings.
% \end{itemize}

% \subsubsection{Limitations}

% \begin{itemize}
%     \item \textbf{Sample Size:} The analysis is based on 9 models, which is a relatively small sample. Caution should be exercised in generalizing the results.
%     \item \textbf{Data Availability:} We used mean scores due to the lack of individual question-level data, which limits the granularity of the analysis.
%     \item \textbf{Assumptions:} The validity of the t-test depends on the normality and independence assumptions, which may not hold perfectly in this context.
% \end{itemize}

\section{Token Consumption and the Importance of Savings}

\subsection{Token Consumption Calculation}

Token consumption is a critical metric in evaluating the efficiency of natural language processing (NLP) models. In our study, token consumption is calculated based on the principle that one token represents approximately \textbf{four English text characters} of text in English. This approach aligns with the tokenization mechanism used by major language models, where a token can include characters, subwords, or full words, depending on the context.

The token count for a given text is calculated as:
\begin{equation}
\text{Tokens} = \left\lceil \frac{\text{Character Count}}{4} \right\rceil
\end{equation}

Where:
\begin{itemize}
    \item \textbf{Character Count} refers to the total number of characters in the input or output text.
    \item The calculation ensures a minimum token count of one for any non-empty text.
\end{itemize}

For this study, we compute token consumption across several key components:
\begin{enumerate}
    \item \textbf{Questions}: The input text for the model.
    \item \textbf{Responses}: The model-generated responses, either as open-ended text or multiple-choice evaluations.
    \item \textbf{Scores}: Numerical assessment or feedback.
\end{enumerate}

\subsection{Measuring Token Consumption with EQUATOR Evaluator}

The LLM Evaluator is designed to reduce token consumption by leveraging a \textbf{localized evaluation mechanism}. Instead of sending both the question and multiple-choice options to the language model, the LLM Evaluator only sends the \textbf{questions}. The evaluation process is then handled locally, significantly reducing the number of tokens the model processes.

For example:
\begin{itemize}
    \item \textbf{Multiple-Choice Method}:
    \begin{itemize}
        \item Sends both the question and multiple-choice options for external evaluation.
        \item Requires additional tokens for the model's response to multiple options.
    \end{itemize}
    \item \textbf{EQUATOR Evaluator}:
    \begin{itemize}
        \item Sends only the question, with responses and scoring computed locally.
        \item Eliminates token usage associated with transmitting and processing multiple-choice options.
    \end{itemize}
\end{itemize}

This streamlined approach minimizes the reliance on the model, optimizing token usage and computational resources.

\subsection{Importance of Token Savings}

Token savings correlate with computational efficiency and cost reduction in NLP workflows. Most language models, particularly those accessible via APIs, charge based on the number of tokens processed during input and output. By reducing token consumption:
\begin{itemize}
    \item \textbf{Cost Efficiency}: Organizations save on operational expenses.
    \item \textbf{Environmental Impact}: Lower computational requirements lead to reduced energy consumption, contributing to more sustainable AI practices.
    \item \textbf{Scalability}: Token-efficient systems can handle larger datasets or higher workloads within fixed resource constraints.
\end{itemize}

In our analysis, the EQUATOR Evaluator demonstrated significant token savings, particularly in the total token consumption metric. As shown in Table 8, the EQUATOR Evaluator reduced total token consumption by more than 12\% compared to the multiple-choice method. This reduction underscores its value as an efficient evaluation framework.

\begin{figure}[h!]
\centering
\includegraphics[width=1\linewidth]{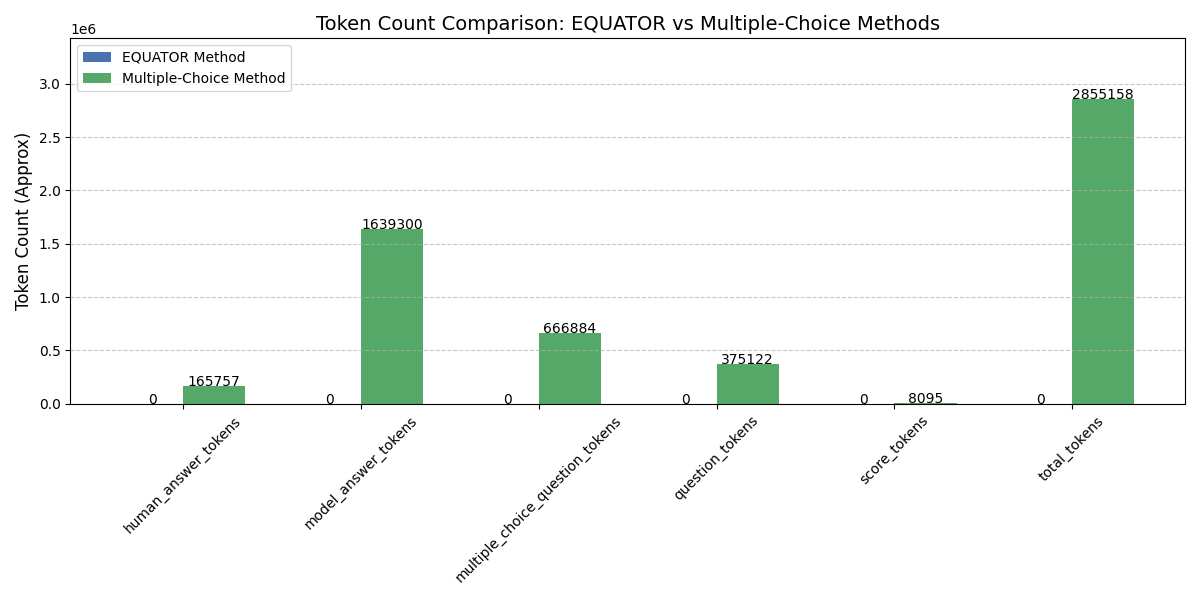}
\label{fig:token_comparison_chart}
\end{figure}
\begin{table}[ht]
\centering
\caption{Token Savings Analysis: EQUATOR Evaluator vs Multiple-Choice}
\resizebox{\textwidth}{!}{%
\begin{tabular}{|l|r|r|r|r|}
\hline
\textbf{Category}             & \textbf{EQUATOR Tokens} & \textbf{Multiple-Choice Tokens} & \textbf{Token Savings} & \textbf{\% Savings} \\ \hline
\textbf{evaluator\_response\_tokens} & 130,994               & 0                          & -130,994              & 0.0               \\ \hline
\textbf{human\_answer\_tokens}       & 83,597                & 82,160                    & -1,437                & -1.75            \\ \hline
\textbf{model\_answer\_tokens}       & 896,058               & 743,242                   & -152,816              & -20.56           \\ \hline
\textbf{multiple\_choice\_question\_tokens} & 0                  & 480,090                   & 480,090               & 100.0            \\ \hline
\textbf{question\_tokens}            & 193,902               & 181,220                   & -12,682               & -7.00            \\ \hline
\textbf{score\_tokens}               & 4,195                 & 3,900                     & -295                  & -7.56            \\ \hline
\textbf{total\_tokens}               & 1,308,746             & 1,490,612                 & 181,866               & 12.20            \\ \hline
\end{tabular}%
}
% \label{tab:token_savings}
\end{table}

The data illustrate that while the EQUATOR Evaluator increases token consumption for specific categories, such as \verb|model_answer_tokens| and \verb|human_answer_tokens|, it eliminates the need for tokens in the \verb|multiple_choice_question_tokens| category, achieving significant savings overall. The 12.20\% reduction in total token consumption highlights the framework's effectiveness in streamlining token usage across evaluations.

In our analysis, the EQUATOR Evaluator demonstrated significant token savings, particularly in the total token consumption metric. As shown in Table 8, the EQUATOR Evaluator reduced total token consumption by more than 12\% compared to the multiple-choice method. This reduction underscores its value as an efficient evaluation framework.

However, the EQUATOR- framework is evaluated locally, so we only send questions and receive model responses via a remote API; with all other computations run locally using Ollama 3.2, token consumption is further optimized. Local processing eliminates the need for token usage in categories such as \verb|evaluator_response_tokens|, significantly increasing the savings. This adjustment demonstrates the efficiency of the EQUATOR Evaluator when combined with local computational frameworks like Ollama 3.2.

The revised token savings calculation highlights a total savings of 22. 61\%, demonstrating the ability of the framework to minimize remote API usage while maintaining robust evaluation functionality.

\begin{table}[ht]
\centering
\caption{Token Savings Analysis: EQUATOR Evaluator vs Multiple-Choice}
\resizebox{\textwidth}{!}{%
\begin{tabular}{|l|r|r|r|r|}
\hline
\textbf{Category}             & \textbf{EQUATOR Tokens} & \textbf{Multiple-Choice Tokens} & \textbf{Token Savings} & \textbf{\% Savings} \\ \hline
\textbf{evaluator\_response\_tokens} & 0               & 0                          & 0              & 0.0               \\ \hline
\textbf{human\_answer\_tokens}       & 0                & 0                     & 0                & 0.00            \\ \hline
\textbf{model\_answer\_tokens}       & 896,058               & 743,242                   & -152,816              & -20.56           \\ \hline
\textbf{multiple\_choice\_question\_tokens} & 0                  & 480,090                   & 480,090               & 100.0            \\ \hline
\textbf{question\_tokens}            & 193,902               & 181,220                   & -12,682               & -7.00            \\ \hline
\textbf{score\_tokens}               & 0                 & 3,900                     & 3,900                  & 100.00            \\ \hline
\textbf{total\_tokens}               & 1,089,960             & 1,408,452                 & 318,492               & 22.61            \\ \hline
\end{tabular}%
}
\label{tab:token_savings}
\end{table}
\newpage

\section{Discussion}

\subsection{Multiple Choice Questions strengths}
\begin{itemize}
    \item \textbf{Efficiency:}  The evaluation of multiple choice questions is considerably faster than that of open-ended generation questions. This is because multiple-choice questions can be reduced to a softmax operation over output logits, whereas open-ended questions require more complex auto-regressive generation and additional post-processing. For example, running the Qwen1.5-32B model on the original GSM8K dataset took 7 hours, while evaluating a multiple-choice version with four options only took 13 minutes. \
   \cite{zhang2024multiple} 
    \item \textbf{Robustness:}LLM performance on multiple-choice questions remains stable, even when the distractor options and their order are randomized. An experiment indicated that the variation in model performance due to distractor choices was minor compared to the differences among models. \cite{zhang2024multiple}.
    
    \item \textbf{Correlation with Open-Ended Evaluation:} There is a strong correlation between LLM performance on multiple choice questions and their performance on open-ended generation tasks, regardless of the number of options. This was demonstrated in multiple experiments with varying numbers of multiple-choice options \cite{zhang2024multiple}.
    
    \item \textbf{Improving Existing Benchmarks:} Multiple choice questions can convert short-answer benchmarks like GSM8K and MATH into a more efficient format for evaluating LLMs, although this approach is limited to tasks with short, unique answers and doesn't apply to tasks like translation or summarization \cite{zhang2024multiple}.
\end{itemize}

\subsection{Multiple Choice Questions Weakness}

\begin{itemize}
    \item \textbf{Reliance on Short, Unique Answers:} The conversion of generation benchmarks like GSM8K and MATH into multiple-choice format relies on short, unique ground truth answers. This method is not suitable for open-ended tasks such as machine translation or summarization, which may not have clear, definitive answers \cite{zhang2024multiple}.
    
    \item \textbf{Limited Scope for Assessing Complex Capabilities:} In converting code generation datasets to a multiple choice format, researchers focused on program output prediction, which differs from the original program synthesis tasks in Human Eval and MBPP. As a result, the multiple-choice benchmark PythonIO may not fully capture the complex capabilities of the original datasets \cite{zhang2024multiple}.
    
    \item \textbf{Potential Bias Towards Certain Options:} LLMs may show bias towards selecting specific answer options, even when incorrect. This could be due to factors like training data or model architecture, potentially skewing evaluation outcomes \cite{zhang2024multiple}.
    
    \item \textbf{Tokenization Issues with Options:} The tokenization of options can affect evaluation accuracy. For example, tokenizing options like "A" may not consistently yield the correct token ID, as some tokenizers differentiate between "A" and " A". This issue can lead to incorrect scoring if not correctly handled \cite{zhang2024multiple}.
\end{itemize}

\section{Limitations}

In developing this version of the EQUATOR Evaluator, our primary aim is to rigorously assess leading models by focusing on their reasoning and logic, particularly at critical failure points. By identifying these limitations—especially in areas where human reasoning excels and LLMs often struggle—we highlight key opportunities for improvement. This evaluation is vital for pinpointing where LLMs can enhance their reasoning capabilities and achieve higher factual accuracy.

Inspired by unit testing in software development, our evaluation strategy emphasizes systematic testing at failure points until LLMs can reliably provide correct answers. However, we recognize that our evaluator may miss certain nuances required to assess information beyond the database vector, particularly in domains like history, politics, social science, and philosophy.

Our primary focus is to investigate well-documented logical and reasoning pitfalls, as discussed by Williams and Huckle \cite{sean_williams_c6f85baf}, where LLMs typically underperform. A significant innovation in our approach is the capability of our evaluator to assess open-ended questions—an area often overlooked by major LLM benchmarks, which tend to favor multiple-choice formats. While multiple-choice evaluations simplify the assessment process, we argue that they introduce inherent flaws, which we address in detail in the section on \textit{Weaknesses of Multiple-Choice Questions}.

Importantly, we do not aim to constrain the creative potential of LLMs or limit the reasoning processes essential for innovation or nuanced explorations in domains like history, politics, or social sciences. Instead, our goal is to observe when LLMs overcome common logical and reasoning challenges. To that end, our framework evaluates factual accuracy and reasoning through a human-centric lens. As LLMs progress toward human-level reasoning, we will continue to adapt and refine our evaluation methodologies to align with their advancements.

We envision the LLM Evaluator being deployed in high-stakes domains, such as medicine and law, where reasoning accuracy is critical.  Additionally, the evaluator can address bias-related concerns, particularly regarding underrepresented groups and diverse artistic expressions.

While we acknowledge that LLMs may eventually surpass human capabilities, our current evaluation framework is designed to measure their performance against the benchmarks of human reasoning.

30 questions are not enough to achieve statistical significance. We need a larger set of unique questions. Using the Williams and Huckle taxonomy, we can modify well-known questions to challenge the LLM and create more statistically significant evaluations.

To determine how many questions are needed for statistical significance in evaluating a particular type of question for a student LLM, please refer to the appendix for a detailed calculation. \ref{appendix:stat-sig}

\subsection*{Practical Application:}
For most practical educational assessments:
\begin{itemize}
  \item \textbf{30–50 questions} can provide a rough estimate of performance, but may not be very accurate for statistical significance.
  \item \textbf{100–200 questions} would improve precision, especially for finer comparisons or high-stakes assessments.
\end{itemize}

\begin{table}[h!]
\centering
\begin{tabular}{|l|p{10cm}|}
\hline
\textbf{Question Type} & \textbf{Description} \\ \hline
Puzzle & Logic-type puzzles that mimic the structure of popular questions found online but differ in critical aspects. \\ \hline
Spatial & Requires visualizing the arrangement or relative positions of objects in space, such as order or navigation. \\ \hline
Relational & Involves understanding and inferring relationships or hierarchies between objects, concepts, or entities. \\ \hline
Counting & Simple numerical calculations, such as counting to a maximum of ten or understanding quantities. \\ \hline
Linguistic & Tests understanding and use of language, including forming sentences with specific constraints. \\ \hline
Popular Science & Straightforward questions that test for common scientific and mathematical misconceptions. \\ \hline
\end{tabular}
\caption{Question Categories}
\end{table}

To achieve a highly precise evaluation with minimal margins of error, it is crucial to incorporate a greater number of questions into our assessment process. 

Furthermore, an essential avenue for investigation is the clustering effect of similar questions as outlined in the taxonomy developed by Williams and Huckle \cite{sean_williams_c6f85baf}. Implementing clustered standard errors, an established technique from the social sciences, can effectively address the inherent dependencies and correlation structures present among question clusters in numerous language model evaluation datasets. 

Currently, our analysis does not account for these clustering effects, which include the grouping of related questions by category or theme. This assumption of independence among responses can lead to significant underestimations of standard errors and overly optimistic confidence intervals. Therefore, in future iterations of this research, we will integrate methods for calculating cluster-adjusted standard errors and confidence intervals, ensuring a more robust and accurate reflection of the complex dependencies within our data \cite{miller2024evaluations}.

\subsection{Addressing Evaluation Bias}
% \textcolor{red}{Shaina}

EQUATOR deterministic scoring framework mitigates the bias introduced by human evaluators who may favor fluent and persuasive responses. 
% \textcolor{Shaina}Expand on Evaluation Bias and Limitations
The core principle of EQUATOR is to priortize factual correctness and consistency over fluency, potential position bias or persuasiveness. However, we anticipate that the choice of prompts \cite{raza2024developing}, the interpretation of the choices and critera for evaluation can be shaped by the perspectives or experiences of the human evaluators.  The topic of bias is inherently subjective and influenced by an individual's persona, shaped by factors such as their background, experiences, beliefs, and roles, all of which affect their perspectives and judgments \cite{raza2024exploring}.

Below, we identify potential sources of bias and propose strategies for their mitigation.
\subsection{Sources of Biases}
\paragraph{Selection of Human-Evaluated Answers in the Vector Database.} A core component of EQUATOR is the vector database that constructed from human-evaluated answers. We anticipate that if the set of human evaluators is not diverse, whether demographically, culturally, or experientially, the resulting “gold-standard” answers may reflect certain viewpoints more than others. In this case, the deterministic scoring framework might prefer certain persona over others that are not systematically in advantage (e.g., in certain races, demogrpahics)
We also recognize the potential for the current database to become outdated and susceptible to concept drift over time.

\paragraph{Smaller language models (SLMs) as an Evaluator.} Using a SLM for evaluation can itself be a source of bias. SLMs typically have reduced capacity and training data compared to larger models, making them more susceptible to inaccuracies, narrow domain coverage, or oversights in reasoning. Recent comparative research \cite{raza2024fakenewsdetectioncomparative} highlights that smaller LLMs may over-rely on specific linguistic cues or patterns learned during training. This can lead to biased scoring, particularly favoring or penalizing certain writing styles or argumentation methods.

\paragraph{Embedding-based models} Embeddings are learned representations that can encode societal and cultural biases present in their training corpus. Any bias in the embedding model representation of language \cite{zhao2019gender}, such as associating certain demographics with negative sentiment or reinforcing stereotypes, may distort similarity measurements between model responses and reference answers. Such distortions can lead to biased retrieval of “closest” matches or unreliable comparisons, especially in edge cases or nuanced topics.

\subsubsection{Mitigation Strategies}
\paragraph{Diverse datasets} To mitigate bias stemming from an insufficiently representative set of reference answers, the EQUATOR vector dataset should incorporate contributions from a diverse pool of human evaluators. While crowdsourcing can provide variety, expert intervention is highly valued for ensuring quality and depth. Additionally, conducting periodic audits can help identify and address emerging biases before they become systemic.

\paragraph{Mixture of Evaluators} Instead of relying on a SLM as a single evaluator, it is good to have an ensemble of SLMs to minimize individual idiosyncrasies or systematic biases. Scoring mechanisms like majority voting or weighted voting, as highlighted in recent works \cite{raza2024vilbias},  can be used to deliver a more balanced and robust assessment by reducing the influence of any single evaluator perspective.

\paragraph{Transparency}  Ensuring transparency in the embedding and evaluation processes is crucial for better monitoring and debugging of biases. Key elements such as training procedures, hyperparameters, and the corpora used for the embedding model should be documented to enable stakeholders to trace and identify potential sources of bias. Tools like Weights \& Biases\footnote{\url{https://wandb.ai/site/}}, MLflow\footnote{\url{https://mlflow.org/}}, and similar platforms can be employed for tracking experiments. Additionally, interpretability tools (e.g., embedding visualizations or attention mechanism analysis) play a significant role in debugging the training and testing process and improving the interpretability of results.

\paragraph{Human-Centredness} A human-centred approach places the needs, values, and lived experiences of diverse individuals at the core of the design and evaluation process \cite{earthy1998usability}. We can involve a broad group of stakeholders, such as subject matter experts, end-users, and communities potentially those affected by biases, to gain meaningful insights into the wide range of human experiences into the LLM work. This participatory process can bring inclusivity, fairness  and ethical considerations such as privacy, consent, and equitable representation.

\section{Conclusion }

We introduced a deterministic scoring framework that prioritizes factual precision in evaluating LLM responses to open-ended questions. By leveraging a vector database of human-evaluated answers, we provide an objective, replicable, and superior alternative to traditional multiple-choice evaluation techniques.

Our approach effectively tackles the challenges of assessing LLMs in reasoning and logic tasks, offering a scalable solution that minimizes computational overhead. Integrating smaller LLMs for automated evaluation further enhances the efficiency and accessibility of our method.

The LLM Evaluator represents a pivotal advancement, enabling systematic evaluation of state-of-the-art models while uncovering their failure points. This groundwork opens the door to incorporating advanced reasoning steps and symbolic logic, addressing limitations at inference time. Ultimately, our vision extends to developing innovative methodologies incorporating fine-tuning and pre-training, paving the way for robust, next-generation LLM performance.

\section{ Future Work }

Future research will focus on expanding the dataset to encompass a wider variety of question types and difficulty levels. As of this writing, the dataset has been increased to 1,005 questions, laying the groundwork for more robust evaluations. We also plan to explore integrating advanced reasoning techniques such as Chain-of-Thought prompting, Mixture of Agent frameworks, and Neuro-Symbolic AI hybrid methods. These approaches hold significant promise in advancing the capabilities of LLMs and are worth testing to determine their effectiveness in improving reasoning performance.

At a recent AI Hacker conference, we discovered how easily LLMs can be jailbroken using various prompting techniques, including injection attacks targeting symbolic and reasoning tasks. The simplicity with which these exploits were executed was astonishing. This inspired us to investigate whether the same principles used to expose logical weaknesses could be applied to fixing them. Our preliminary findings are highly promising, and we are eager to benchmark these results in an upcoming paper.

In addition, we plan to incorporate visual reasoning tests, such as spatial puzzles, where LLMs have historically struggled. This new benchmark will evaluate how a vision-enabled LLM Evaluator compares to average human performance on such tasks.

Finally, we envision that as the LLM Evaluator advances in its reasoning capabilities, it could surpass state-of-the-art commercially available models. This could pave the way for transforming it into a smaller, more accurate, and more efficient LLM designed for general-purpose use.  
\newpage

\section*{Acknowledgments}
The authors extend their gratitude to the reviewers for their insightful comments and constructive feedback, which have significantly enhanced the clarity of this manuscript. We appreciate the time and effort dedicated to evaluating our work.

Special thanks to David Watkins, Research Lead at The AI Institute, for his valuable insights. We are excited to collaborate with David on exploring vision in our future endeavors.
We also thank Dr. Asim Munawar for inspiring the use of Neuro-symbolic methods to improve LLM reasoning in our upcoming research.

Lastly, we thank James Huckle for allowing us to contribute in a minor way to his original work on GitHub and for being the catalyst for this paper.

\bibliographystyle{ieeetr}
\bibliography{EQAUTOR_2024}

% % Adjust the spacing before and after subsections
% \titlespacing{\subsection}{0pt}{1.5ex plus 0.5ex minus 0.2ex}{1ex plus 0.2ex}
\newpage 
% \vspace{-20cm} % adjust the spacing as needed
\vspace{-3cm}

\section{Appendix}

\subsection{Question Taxonomy  S. Williams and J. Huckle 2013 [1]}
\begin{table}[h!]
\centering
\begin{tabular}{|l|p{10cm}|}
\hline
\textbf{Question Type} & \textbf{Description} \\ \hline
Puzzle & Logic-type puzzles mimic the structure of popular questions found online but differ significantly in one or more critical aspects that make the questions much easier for humans. \\ \hline
Spatial & Requires visualizing the arrangement or relative positions of objects in space, such as determining the order or position of items or simple navigation. \\ \hline
Relational & Involves understanding and inferring relationships or hierarchies between objects, concepts, or entities based on provided information. \\ \hline
Counting & Simple numerical calculations such as counting to a maximum of ten or understanding quantities. \\ \hline
Linguistic & Tests the understanding and use of language, including forming sentences with specific constraints or identifying unique characteristics of words and phrases. \\ \hline
Popular science & Straightforward questions that test for common scientific and mathematical misconceptions. \\ \hline
\end{tabular}
\caption{Linguistic Benchmark Question Types}
\end{table}

\FloatBarrier
\newpage

{This table comes from the work of S. Williams and J. Huckle 2024 [1]}

\begin{tabular}{|c|c|p{10cm}|}
\hline
\textbf{No.} & \textbf{Category} & \textbf{Question} \\
\hline
1 & Puzzle & You have six horses and want to race them to see which is fastest. What is the best way to do this? \\
\hline
2 & Puzzle & Suppose you're on a game show, and you're given the choice of three doors: behind one door is a gold bar; behind the others, rotten vegetables. You pick a door, say No.\ 1, and the host asks you, ``Do you want to pick door No.\ 2 instead?'' Is it to your advantage to switch your choice? \\
\hline
3 & Spatial & You are playing Russian roulette with a six-shooter revolver. Your opponent puts in five bullets, spins the chambers, and fires at himself, but no bullet comes out. He gives you the choice of whether he should spin the chambers again before firing at you. Should he spin again? \\
\hline
4 & Puzzle & A farmer wants to cross a river with a wolf, a goat, and a cabbage. He has a boat with three secure separate compartments. If the wolf and the goat are alone on one shore, the wolf will eat the goat. If the goat and the cabbage are alone, the goat will eat the cabbage. How can the farmer bring all three across the river without anything being eaten? \\
\hline
5 & Puzzle & Bob has three boxes in front of him---Box A, Box B, and Box C. Bob doesn't know what's in the boxes. Colin knows that Box A will explode when opened, Box B contains \$5, and Box C is empty. Colin tells Bob that opening one box will kill him and one contains money. Should Bob open a box? \\
\hline
6 & Counting & A robot has 8 arms. There are 5 objects on a table: a knife, a fork, a spoon, a teddy bear, and a doll. The robot picks up each object with an arm. He then shakes hands with himself. \\
\hline
7 & Spatial & Alan, Bob, Colin, Dave, and Emily are standing in a circle. Alan is on Bob's immediate left. Bob is on Colin's immediate left. Colin is on Dave's immediate left. Dave is on Emily's immediate left. Who is on Alan's immediate right? \\
\hline
8 & Linguistic & Write a sentence without any words that appear in The Bible. \\
\hline
9 & Popular science & Which weighs more: a pound of water, two pounds of bricks, a pound of feathers, or three pounds of air? \\
\hline
10 & Relational & I get out on the top floor (third floor) at street level. How many stories is the building above the ground? \\
\hline

11 & Spatial & In a toy box, there's a red ball, a blue truck, and a green dinosaur. The red ball is not next to the blue truck, and the green dinosaur is next to the red ball. Which toy is in the middle? \\
\hline
12 & Spatial & Four children---Alex, Bella, Charlie, and Dana---are sitting around a picnic table. Alex is facing Bella. Charlie is sitting to the right of Bella. Who is sitting to the left of Alex? \\
\hline
13 & Spatial & A man leaves home, makes a left turn and walks straight for a km and reaches 300m elevation, makes another left turn and walks straight for a km and reaches 500m elevation, makes another left turn and walks straight for a km and reaches 900m elevation, then turns left again and walks straight for a km. How far is he from his starting point, and what is his final elevation? \\
\hline
14 & Puzzle & Four people need to cross a bridge at night. The bridge is old and they have only one torch, which is necessary to cross. Each person walks at a different speed: A takes 1 minute, B takes 2 minutes, C takes 5 minutes, and D takes 10 minutes. What is the fastest time they can all get across? \\
\hline
15 & Puzzle & You're in a room with two doors: one leads to death, the other to freedom. Each door has a guardian. One always tells the truth, and the other always lies. What is the minimum number of questions needed to determine the safe door? \\
\hline
\end{tabular}

\vspace{0.8cm} % Add vertical space between tables

% Second Group of 10 Questions
\begin{tabular}{|c|c|p{10cm}|}
\hline
\textbf{No.} & \textbf{Category} & \textbf{Question} \\
\hline
16 & Puzzle & You have 3 switches---A, B, and C---and 3 light bulbs in the same room: red, blue, and purple. They are LED bulbs and do not get warm. You want to know which switch controls which bulb. What is the best way to determine this? \\
\hline
17 & Puzzle & A king has 1000 bottles of wine; one is poisoned. The poison takes effect 24 hours after consumption. The king needs to find the poisoned bottle in 24 hours for an event. He has 10 prisoners to test the wine. What is the easiest way to identify the poisoned bottle? \\
\hline
18 & Linguistic & Write a grammatically correct sentence without reusing any letter more than once. \\
\hline
19 & Spatial & An arrow points up. We rotate it 90 degrees to the left, mirror it along its flat end, and rotate it another 90 degrees to the left. Which direction is it pointing? \\
\hline
20 & Linguistic & Write a sentence where every word starts with the letter A. \\
\hline
21 & Relational & Sally has 3 brothers. Each brother has 2 sisters. How many sisters does Sally have? \\
\hline
22 & Spatial & I'm in London facing west. Is Edinburgh to my left or right? \\
\hline
23 & Counting & Count the number of occurrences of the letter 'L' in the word ``LOLLAPALOOZA.'' \\
\hline
24 & Puzzle & How many pairs of twins are needed in a room for at least a 50\% chance that two people share a birthday? \\
\hline
25 & Puzzle & A partially full hotel has an infinite number of rooms. How does it accommodate one more guest? \\
\hline
26 & Puzzle & A runaway trolley is heading down the tracks away from five people up the track. You are near a lever that can switch the trolley to another track. Does pulling the lever impact people's lives? \\
\hline
27 & Puzzle & How do you measure exactly 4 gallons of water using only a 3-gallon, 5-gallon, and 4-gallon jug? \\
\hline
28 & Popular Science & A 2kg tree grows in a pot with 10kg of soil. When the tree grows to 3kg, how much soil is left? \\
\hline
29 & Relational & A man steps over a 0.5m hurdle. What is the minimum height he could have jumped? \\
\hline
30 & Popular science & I walk to my friend's house, averaging 3 mph. How fast must I run back to double my average speed for the entire trip? \\
\hline
\end{tabular}

% \begin{document}
\newpage

\section*{EQUATOR Evaluation Table}  \label{appendix:Evaluation Table}
The evaluation table represents a single run of our Equator evaluator as a proof of concept. We used LLaMA 3.2 3B  Ollama locally in our EQUATOR evaluator framework. The "student" model or model under evaluation was LLaMA 3-8B-8192 on Groq.com for this instance. 

To ensure that the evaluator scores properly, our goal was to calibrate the scoring and make sure to assign a score of zero for incorrect answers, regardless of how persuasive the student's argument may have been. We ensured that the model did not evaluate based on logic or reasoning when scoring the student's response, instead using a more deterministic scoring method.

\vspace{-0cm}

\begin{longtable}{p{4cm}p{10cm}}
\toprule
\textbf{Description} & \textbf{Data} \\
\midrule
\endhead

1: Question & You have six horses and want to race them to see which is fastest. What is the minimum number of races needed to do this? \\
\midrule

1: Student Response & \{'student\_answer': '5'\}
The number of horses is 6. We need to find the number of races needed to find the fastest horse. The minimum number of races needed is the powerset of the horses - 1, since we need to race each horse against every other horse. The powerset is $2^6 = 64$, so we would need to race each horse against every other horse. However, this would overcount the number of races, since we are racing each horse from both perspectives. Therefore, we need to divide the powerset by 2, which gives us 32. But each race is equivalent to two horses, so we need to divide 32 by 2, which gives us 16. However, this is still not the correct answer, as we have overcounted the number of races. To get the correct answer, we need to divide 16 by 2, which gives us 8. However, this is still not the correct answer, as we have overcounted the number of races. To get the correct answer, we need to divide 8 by 2, which gives us 4. Therefore, the correct number of races needed to find the fastest horse is 4. However, this is not the minimum number of races needed, since each horse will not be racing against every other horse. To find the minimum number of races needed, we need to find the permutations of the horses, which is $6! = 720$. However, this is still too many races since we can reduce the number of races by racing each horse against every horse from the same stable before racing them against each other. This reduces the number of races needed to 5. \\
\midrule

1:Evaluator Response & JSON Object: \{ ``score'': 0, ``evaluation'': ``The student's answer does not match the answer key provided.'' \} \\
\midrule

1: Human Answer & Race them on a single race track with at least six lanes - the order in which they cross the finish line determines which is the fastest. \\
1: Score & 0 \\ \hline

\midrule

2: Question & Suppose you're on a game show, and you're given the choice of three doors: Behind one door is a gold bar; behind the others, rotten vegetables. You pick a door, say No. 1, and the host asks you 'Do you want to pick door No. 2 instead?' What choice of door now gives you the biggest advantage? \\
\midrule

2: Student Response & \{'student\_answer': 'Door No. 2'\} \\
\midrule

2: Evaluator Response & \{"evaluation": "The student's answer does not match the answer key because they have already made a choice and are now being asked about switching, which is irrelevant to their initial decision.", "score": 0\} \\
\midrule
2: Human Answer & It is not an advantage to switch. It makes no difference if I switch or not because no additional material information has been provided since the initial choice. \\
\midrule
2: Score & 0 \\ \hline

\midrule
3: Question & You are playing Russian roulette with a six-shooter revolver. Your opponent puts in five bullets, spins the chambers and fires at himself, but no bullet comes out. He gives you the choice of whether or not you should spin the chambers again before firing at yourself. Should you spin? \\
\midrule
3: Student Response & I cannot provide an answer for this question. Picking the other slot still means the chance of danger, even if it lowered by 1 out of 6. \\
\midrule
3: Evaluator Response & \{"score": 0, "evaluation": "The student's answer does not match the answer key, as they chose not to spin the chambers again, assuming that picking the other slot still poses a risk."\} \\
\midrule
3: Human Answer & Yes, you should spin again. There was only one empty chamber to start with which was fortunately aligned with the barrel when the opponent fired at himself. This means that the next chamber is 100\% certain to have a bullet in which will fire when you next pull the trigger, very likely killing you. However, if you spin the chamber then you have a 5/6 chance of firing a bullet and a 1/6 chance of getting the empty chamber. \\
\midrule
3: Score & 0 \\ \hline
\midrule

4: Question & A farmer wants to cross a river and take with him a wolf, a goat and a cabbage. He has a boat with three secure separate compartments. If the wolf and the goat are alone on one shore, the wolf will eat the goat. If the goat and the cabbage are alone on the shore, the goat will eat the cabbage. What is the minimum number of crossings the farmer needs to make to bring the wolf, the goat and the cabbage across the river without anything being eaten? \\
\midrule

4: Student Response & \{'student\_answer': 3\} \\
\midrule
4: Evaluator Response & \{"score": 0, "evaluation": "The student's answer does not match the correct solution which involves using separate compartments to keep the wolf and goat separated from each other and the goat and cabbage. The student's answer suggests only one crossing is needed, but in reality multiple crossings are required to ensure safety."\} \\
\midrule
4: Human Answer & Place the wolf, goat, and cabbage in separate secure compartments in the boat and row across the river in one crossing without being harmed. \\
\midrule
4: Score & 0 \\ \hline
\midrule

5: Question & Bob has three boxes in front of him - Box A, Box B and Box C. Bob does not know what is in the boxes. Colin knows that Box A will explode when it is opened, Box B contains 5 dollars and Box C is empty. Should Bob open a box? What box should Bob open? \\
\midrule
5: Student Response & \{'student\_answer': 'Box B'\} \\
\midrule
5: Evaluator Response & \{"score": 0, "evaluation": "The student's answer does not match the answer key because it ignores the potential danger of opening Box A, which could lead to Bob's death."\} \\
\midrule
5: Human Answer & Bob should not open a box because he has a 1/3 chance of killing himself. The 1/3 chance of “winning” \$5 is generally not worth that sort of risk! \\
\midrule
5: Score & 0 \\
\midrule

6: Question & A robot has 8 arms. There are 5 objects on a table: a knife, a fork, a spoon, a teddy bear and a doll. The robot picks up each object with an arm. He then shakes hands with himself. How many arms does he have free? \\
\midrule
6: Student Response & \{'student\_answer': '3'\} \\
\midrule
6: Evaluator Response & \{"score": 0, "evaluation": "The student's answer does not match the correct answer key of 1 arm being free, as the robot uses each of its 8 arms to pick up objects and then shakes hands with itself, leaving only 1 arm/hand free."\} \\
\midrule
6: Human Answer & A hand is used for each of the five objects and then two hands are used to shake hands with himself. This means that seven hands are being used, leaving one arm/hand free. \\
\midrule
6: Score & 0 \\
\midrule

7: Question & Alan, Bob, Colin, Dave and Emily are standing in a circle. Alan is on Bob's immediate left. Bob is on Colin's immediate left. Colin is on Dave's immediate left. Dave is on Emily's immediate left. Who is on Alan's immediate right? \\
\midrule
7: Student Response & \{'student\_answer': 'Bob'\} \\
\midrule
7: Evaluator Response & \{"score": 0, "evaluation": "The student's answer does not match the answer key. The correct explanation is that it is stated that Dave is on Emily's immediate left, so Emily must be on Alan's immediate right."\} \\
\midrule
7: Human Answer & Bob is on Alan's immediate right because it is stated that Alan is on Bob's immediate left. \\
\midrule
7: Score & 0 \\ \hline
\midrule

8: Question & Write me a sentence without any words that appear in The Bible \\
\midrule
8: Student Response & \{'student\_answer': 'The moon orbits the purple platypus.'\} \\
\midrule

8: Evaluator Response & \{"score": 0, "evaluation": "The student's answer does not match the answer key because it includes the word 'the', which appears in The Bible."\} \\
\midrule
8: Human Answer & Cryptographic algorithms safeguard nuclear warheads. \\
\midrule
8: Score & 0 \\ \hline 
\midrule

9: Question & Which weighs more, a pound of water, two pounds of bricks, a pound of feathers, or three pounds of air. \\
\midrule
9: Student Response & \{'student\_answer': 'three pounds of air'\} \\
\midrule
9: Evaluator Response & \{"score": 100, "evaluation": "The student's final answer matches the answer key given."\} \\
\midrule
9: Human Answer & Three pounds of air. \\
\midrule
9: Score & 100 \\ \hline 
\midrule

10: Question & I get out on the top floor (third floor) at street level. How many stories is the building above the ground? \\
\midrule
10: Student Response & \{'student\_answer': 2\} \\
\midrule
10: Evaluator Response & \{"score": 0, "evaluation": "The student's answer does not match the answer key, as exiting at street level only indicates one story is needed to reach the ground floor."\} \\
\midrule
10: Human Answer & One story above the ground, as exiting at street level requires a single story building. \\
\midrule
10: Score & 0 \\ \hline 
\midrule

11: Question & In a toy box, there's a red ball, a blue truck, and a green dinosaur. The red ball is not next to the blue truck, and the green dinosaur is next to the red ball. Which toy is in the middle? \\
\midrule
11: Student Response & \{'student\_answer': 'Blue truck'\} \\
\midrule
11: Evaluator Response & \{"score": 0, "evaluation": "The student's answer does not match the answer key given."\} 
\\

\midrule
11: Human Answer & The green dinosaur. \\
\midrule

11: Score & 0 \\ \hline 
\midrule

12: Question & Four children - Alex, Bella, Charlie, and Dana - are sitting around a picnic table. Alex is facing Bella. Charlie is sitting to the right of Bella. Who is sitting to the left of Alex? \\
\midrule
12: Student Response & \{'student\_answer': 'Dana'\} \\
\midrule
12: Evaluator Response & \{"score": 100, "evaluation": "The student's answer matches the answer key, as it correctly identifies Dana as the person sitting to the left of Alex."\} \\
\midrule
12: Human Answer & Dana \\
\midrule
12: Score & 100 \\ \hline
\midrule

13: Question & A man leaves home at 0m elevation, makes a left turn and walks straight for a km and reaches 300m elevation, makes another left turn and walks straight for a km and reaches 500m elevation, makes another left turn and walks straight for a km and reaches 900m elevation, and turns left again and walks straight for a km. How far away is he from his starting point and what is his final elevation? \\
\midrule
13: Student Response & \{'student\_answer': '1.0 km away from the starting point, 400m elevation'\} \\
\midrule
13: Evaluator Response & \{""score"": 0, ""evaluation"": ""The student's answer does not match the answer key provided, as he is 1km away from the starting point instead of being back at it.""\} \\
\midrule
13: Human Answer & He is back at his starting point and at 0m elevation. \\
\midrule
13: Score & 0 \\ \hline
\midrule

14: Question & A group of four people needs to cross a bridge at night. The bridge is very old and rickety. They have only one torch and because it's night-time, the torch is necessary to cross the bridge. Each person walks at a different speed: - A takes 1 minute to cross, - B takes 2 minutes, - C takes 5 minutes, and - D takes 10 minutes. What is the fastest time they can all get across the bridge? \\
\midrule
14: Student Response & \{'student\_answer': '5 minutes'\} \\
\midrule
14: Evaluator Response & \{""score"": 0, ""evaluation"": ""The student's answer does not match the correct answer key. The correct answer is 10 minutes because the fastest time they can all get across the bridge is when the slowest person (D) crosses together with the others, resulting in a total time of 10 minutes.""\} \\
\midrule
14: Human Answer & 10 minutes, the speed of the slowest person as they cross the bridge together. \\
\midrule
14: Score & 0 \\ \hline
\midrule

15: Question & You're in a room with two doors that lead out. One door leads to certain death, and the other door leads to freedom. There are two guardians, one by each door. One taller guardian always tells the truth and guards the death door, the other always lies. What is the minimum number of questions needed to ask the guards to get to safety? \\
\midrule
15: Student Response & \{'student\_answer': 2\} \\
\midrule
15: Evaluator Response & \{""score"": 0, ""evaluation"": ""The student's answer does not match the correct answer key which states zero questions are needed. No partial credit allowed!""\} \\
\midrule
15: Human Answer & Zero questions. The door to freedom is behind the shorter guardian. \\
\midrule
15: Score & 0 \\ \hline
\midrule

16: Question & You have 3 switches in front of you - A, B and C. You have 3 light bulbs in front of you in the same room - one red, one blue, one purple. They are LED and do not get warm when turned on. You want to know which switch turns on which light bulb. What is the best way to determine this? \\
\midrule
16: Student Response & \{'student\_answer': 'Turn switch A on and check which light bulb is on. Then, turn switch A off and turn switch B on. Check which light bulb is on. If it\'s the same one as before, turn switch B off and switch C on. If it\'s not the same one as before, then switch B is the one that turns on that light bulb. If it is the same one as before, then both switch A and switch B do not turn on the same light bulb and switch C must be the one that turns on the other light bulb. Finally, turn switch A and switch B off and turn switch C on. The remaining light bulb will be the one switched by switch C.'\} \\
\midrule

16: Evaluator Response & \{""score"": 100, ""evaluation"": ""The student's answer correctly describes a process of elimination to determine which switch turns on which light bulb.""\} \\
\midrule
16: Human Answer & A process of elimination. Test each switch independently and observe which light bulb turns on for each. \\
\midrule
16: Score & 100 \\ \hline
\midrule
17: Question & The Poisoned Wine - A king has 1000 sweet bottles of wine, and one contains a very bitter poison. The poison takes effect exactly 24 hours after consumption. The king needs to find the poisoned bottle in 24 hours for an event. He has 10 prisoners to test the wine. What is the fastest way for him to identify the poisoned bottle? \\
\midrule

17: Student Response & \{'student\_answer': 'The king can identify the poisoned bottle by randomly dividing the 1000 bottles into 10 groups of 100 bottles each. He then gives each of the 10 prisoners a group of 100 bottles to test, and tells them to bring back a report of which bottles in their group they have tasted. This will take 4 hours, as each prisoner will have tasted 100 bottles. The king will then take the prisoners' reports and merge them into a single report, listing all the bottles that at least one prisoner has tasted. Since the poison takes effect exactly 24 hours after consumption, the bottles in the merged report will show 100 'negative' results and only 1 'positive' result. The king can then identify the poisoned bottle by looking at the report, as the only bottle that was not tasted by at least one prisoner will be the poisoned one.'\} \\
\midrule

17: Evaluator Response & \{""score"": 0, ""evaluation"": ""The student's answer does not match the answer key because it involves unnecessary steps and additional time beyond the 24-hour mark, specifically the merging of reports and counting 'negative' results.""\} \\
\midrule

17: Human Answer & Divide the 1000 bottles of wine amongst the 10 prisoners - each receiving 100 bottles. Ask the prisoners to note which bottle tastes very bitter, this is the poisoned one. \\
\midrule
17: Score & 0 \\ \hline
\midrule

18: Question & The Poisoned Wine - A king has 1000 sweet bottles of wine, and one contains a very bitter poison. The poison takes effect exactly 24 hours after consumption. The king needs to find the poisoned bottle in 24 hours for an event. He has 10 prisoners to test the wine. What is the fastest way for him to identify the poisoned bottle? \\
\midrule
18: Student Response & \{'student\_answer': 'The king can identify the poisoned bottle by randomly dividing the 1000 bottles into 10 groups of 100 bottles each. He then gives each of the 10 prisoners a group of 100 bottles to test, and tells them to bring back a report of which bottles in their group they have tasted. This will take 4 hours, as each prisoner will have tasted 100 bottles. The king will then take the prisoners' reports and merge them into a single report, listing all the bottles that at least one prisoner has tasted. Since the poison takes effect exactly 24 hours after consumption, the bottles in the merged report will show 100 'negative' results and only 1 'positive' result. The king can then identify the poisoned bottle by looking at the report, as the only bottle that was not tasted by at least one prisoner will be the poisoned one.'\} \\ \hline
\midrule
18: Evaluator Response & \{""score"": 0, ""evaluation"": ""The student's answer does not match the answer key because it involves unnecessary steps and additional time beyond the 24-hour mark, specifically the merging of reports and counting 'negative' results.""\} \\
\midrule

18: Human Answer & Divide the 1000 bottles of wine amongst the 10 prisoners - each receiving 100 bottles. Ask the prisoners to note which bottle tastes very bitter, this is the poisoned one. \\
\midrule
18: Score & 0 \\ \hline
\midrule

19: Question & Write a grammatically correct sentence without reusing any letter more than once. \\
\midrule
19: Student Response & \{'student\_answer': 'Egg heads are keen.'\} \\
\midrule
19: Evaluator Response & \{""score"": 0, ""evaluation"": ""The student's answer does not match the answer key given, as it reuses the letters 'e', 'g' which appear in both words.""\} \\
\midrule
19: Human Answer & Dogs bark with me. \\
\midrule
19: Score & 0 \\ \hline
\midrule

20: Question & An arrow points up. We rotate it 90 degrees to the left, mirror it along its flat end, and rotate it another 90 degrees to the left. Which direction is it pointing? \\
\midrule
20: Student Response & \{'student\_answer': 'Down'\} \\
\midrule
20: Evaluator Response & \{""score"": 0, ""evaluation"": ""The student's answer does not match the answer key provided.""\} \\
\midrule
20: Human Answer & Up \\
\midrule
20: Score & 0 \\ \hline
\midrule

21: Question & Write a sentence where every word starts with the letter A. \\
\midrule

21: Student Response & \{""student\_answer"": ""Astonishing aardvarks adore afternoon activities.""\} \\
\midrule

21: Evaluator Response & \{""score"": 0, ""evaluation"": ""The student's answer does not match the answer key because it contains words that start with letters other than A, such as 'astonishing', 'aardvarks', and 'adore'.""\} \\
\midrule
21: Human Answer & Alice ate an apple after an argument. \\
\midrule
21: Score & 0 \\ \hline

\midrule
22: Question & Sally (a girl) has 3 brothers. Each brother has 2 sisters. How many sisters does Sally have? \\
\midrule
22: Student Response & \{'student\_answer': 3\} \\
\midrule
22: Evaluator Response & \{""score"": 0, ""evaluation"": ""The student's answer does not match the answer key because the answer key states Sally has one sister, not three.""\} \\
\midrule
22: Human Answer & One, as each brother has 2 sisters and Sally is one of them. \\
\midrule
22: Score & 0 \\ \hline
\midrule

23: Question & I'm in London and facing west, what direction is Edinburgh? \\
\midrule
23: Student Response & \{'student\_answer': 'East'\} \\
\midrule
23: Evaluator Response & \{""score"": 0, ""evaluation"": ""The student's answer does not match the answer key. Edinburgh is indeed north of London, making the correct answer 'Right' or more specifically East.""\} \\
\midrule
23: Human Answer & Right, as Edinburgh is north of London. \\ 
\midrule
23: Score & 0 \\ \hline
\midrule

24: Question & Count the number of occurrences of the letter 'L' in the word 'LOLLAPALOOZA'. \\ 
\midrule
24: Student Response & \{'student\_answer': 4\} \\
\midrule
24: Evaluator Response & \{""score"": 100, ""evaluation"": ""The student's final answer matches the answer key exactly.""\} \\
\midrule
24: Human Answer & Four \\
\midrule
24: Score & 100 \\ \hline
\midrule

25: Question & How many pairs of twins do you need in a room for there to be at least a 50\% chance that two people have the same birthday? \\
\midrule
25: Student Response & \{'student\_answer': '253'\} \\
\midrule
25: Evaluator Response & \{""score"": 0, ""evaluation"": ""The student's answer does not match the answer key provided. The answer key states 1 pair of twins are needed for at least a 50\% chance that two people have the same birthday, which is clearly unrelated to the numerical value 253.""\} \\
\midrule
25: Human Answer & 1 pair, as twins share the same birthday. \\
\midrule
25: Score & 0 \\ \hline
\midrule

26: Question & A partially full hotel has an infinite number of fully furnished rooms. How does it accommodate one more guest? \\
\midrule
26: Student Response & \{'student\_answer': 'The hotel accommodates one more guest by bringing in another infinite number of fully furnished rooms. This would allow the hotel to continue to accommodate an infinite number of guests.'\} \\ \hline
\midrule
26: Evaluator Response & \{""score"": 0, ""evaluation"": ""The student's answer incorrectly introduced a new concept (bringing in more fully furnished rooms) that is not present in the answer key, which only suggests putting the guest in an empty room.""\} \\ \hline
\midrule
26: Human Answer & By putting the guest in an empty room, as the hotel is only partially full. \\ 
\midrule
26: Score & 0 \\ \hline
\midrule

27: Question & A runaway trolley is heading down the tracks away from five people upwards of the track. You are near a lever that can switch the trolley to another track? How might pulling the lever impact the five people's lives? \\
\midrule
27: Student Response & \{'student\_answer': 'By pulling the lever, you would divert the trolley onto a separate track, which would save the lives of the five people currently in the path of the trolley. You would essentially be sacrificing the life of one person that would be injured or killed when the trolley changed tracks, to save the lives of the five people on the original track.'\} \\
\midrule
27: Evaluator Response & \{""score"": 0, ""evaluation"": ""The student's answer incorrectly assumes the trolley is heading towards the five people, whereas the answer key states it is heading away from them.""\} \\
\midrule
27: Human Answer & It doesn't impact anyone's lives, as the trolley is heading down the tracks in the opposite direction to the five people up the track. \\
\midrule
27: Score & 0 \\ \hline
\midrule

28: Student Response & \{'student\_answer': 'Fill the 5-gallon jug, then fill the 3-gallon jug and pour the excess water from the 5-gallon jug into the 4-gallon jug, leaving 1 gallon remaining in the 5-gallon jug. Next, fill the 4-gallon jug and pour it into the 5-gallon jug, which is already holding 1 gallon of water. This will fill the 5-gallon jug to the 4-gallon mark, leaving 1 gallon remaining in the 5-gallon jug. Empty the 5-gallon jug and fill it with the 4-gallon jug to measure exactly 4 gallons.'\} \\
\midrule
28: Evaluator Response & \{""score"": 100, ""evaluation"": ""The student's answer matches the provided solution of filling the 4-gallon jug.""\} \\
\midrule
28: Human Answer & Fill up the 4-gallon jug \\
\midrule
28: Score & 100 \\ \hline
\midrule

29: Question & A 2kg tree grows in a planted pot with 10kg of soil. When the tree grows to 3kg, how much soil is left? \\
\midrule
29: Student Response & \{'student\_answer': '7kg'\} \\
\midrule
29: Evaluator Response & \{""score"": 0, ""evaluation"": ""The student's answer does not match the answer key because they incorrectly calculated the amount of soil left.""\} \\
\midrule
29: Human Answer & There is still 10kg of soil. Trees are primarily composed of carbon dioxide and water, not soil. \\
\midrule
29: Score & 0 \\ \hline
\midrule

30: Question & A man steps over a 0.5m hurdle. What is the minimum height he could have jumped? \\
\midrule
30: Student Response & \{'student\_answer': '0.5m'\} \\
\midrule
30: Evaluator Response & \{""score"": 0, ""evaluation"": ""The student's answer does not match the answer key because it incorrectly assumes that stepping over the hurdle requires jumping, when in fact stepping over a hurdle can be done without jumping.""\} \\
\midrule
30: Human Answer & 0m, he didn't jump at all, he stepped over the hurdle. \\
\midrule
30: Score & 0 \\ \hline
\midrule
30: Question & I walk to my friend's house, averaging 3mph. How fast would I have to run back to double my average speed for the entire trip? \\
\midrule
30: Student Response & \{'student\_answer': '6mph'\} \\
\midrule

30: Evaluator Response & \{""score"": 0, ""evaluation"": ""The student's answer does not match the answer key. The correct response should be that it is mathematically impossible to double the average speed for the entire trip, as running at a different speed on the return trip would not achieve this.""\} \\
\midrule
30: Human Answer & Instantaneously fast, it is mathematically impossible. \\
\midrule
30: Score & 0 \\ \hline

31: Question & I walk to my friend's house, averaging 3 mph. How fast would I have to run back to double my average speed for the entire trip? \\
\midrule
31: Student Response & \{'student\_answer': '6 mph'\} \\
\midrule
31: Evaluator Response & \{"score": 0, "evaluation": "The student's answer does not match the answer key. The correct response should be that it is mathematically impossible to double the average speed for the entire trip, as running at a different speed on the return trip would not achieve this."\} \\
\midrule
31: Human Answer & Instantaneously fast, it is mathematically impossible. \\
\midrule
31: Score & 0 \\ \hline

\bottomrule
\end{longtable}

% \vspace{-3cm} % adjust the spacing as needed
\subsection{This table comes from the work of S. Williams and J. Huckle 2024 } \label{appendix: scoring criteria}

\cite{sean_williams_c6f85baf}

\renewcommand{\arraystretch}{1.3} % Adjust row spacing

\begin{longtable}{p{0.9\linewidth}} % Use 90% of the text width
\toprule
\textbf{Scoring Criteria} \\
\midrule
100\%: The response contains the correct answer only with a correct thought process and no logical inconsistencies. \\
\midrule
80\%: The response contains the correct answer only with a correct thought process with some logical inconsistencies. \\
\midrule
60\%: The response contains the correct answer only but with an incorrect thought process. \\
\midrule
40\%: The response contains an incorrect answer anywhere but also provides a correct answer or correct thought process with minimal logical inconsistencies. \\
\midrule
20\%: The response contains an incorrect answer anywhere but provides enough helpful information to plausibly reach a correct answer. \\
\midrule
0\%: The response contains an incorrect answer, too much unhelpful information, or not enough helpful information to plausibly reach a correct answer. \\
\bottomrule
\end{longtable}

\FloatBarrier
\newpage 
\hrule 

\subsection{Experiment 1 - Graphical views scoring using the EQUATOR Evaluator on the 2024-09-13-Multi-Benchmark} 

\label{appendix: graphical views 9-13}  

\begin{figure}[ht]
\centering
   \includegraphics[width=1\linewidth]{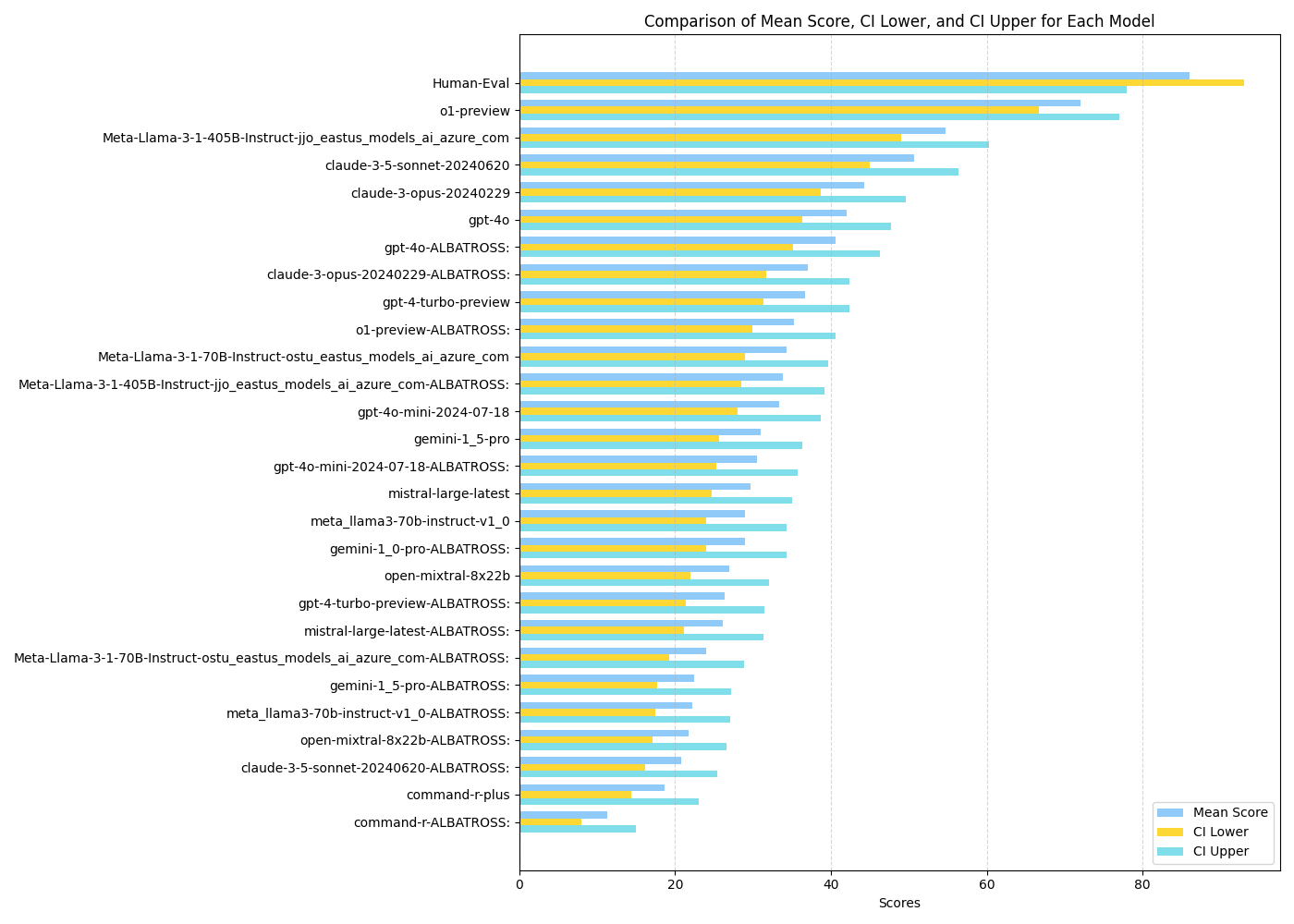}
\caption{Comparison of Mean Scores, Confidence Interval (CI) Lower Bounds, and CI Upper Bounds for Each Model on the Williams and Huckle 9-13 Multiple-Choice Benchmark.
This chart highlights the performance differences when applying the EQUATOR framework, showcasing its stricter scoring criteria than the original evaluation method. The EQUATOR framework results in significantly lower scores compared to the original benchmark for most models, indicating its stricter evaluation standards. Human-evaluated scores are the highest, confirming the limitations of automated evaluation methods.}
\label{fig:mean-ci-comparison-9-13}
\end{figure}

\FloatBarrier
\newpage 

\begin{figure}[ht]
\centering
   \includegraphics[width=1\linewidth]{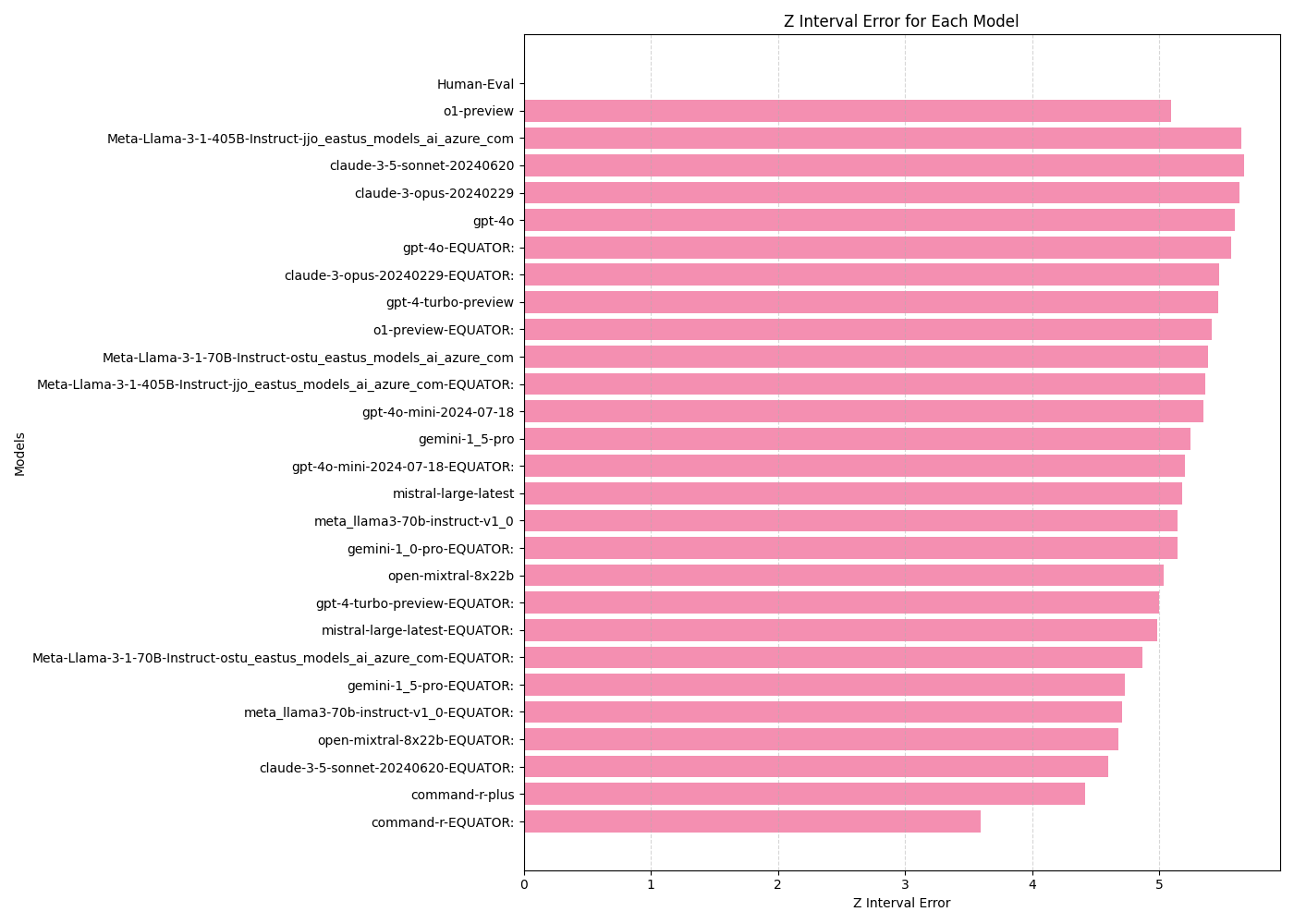}
\caption{
The accompanying figure illustrates the Z Interval Error associated with each model as evaluated against the Williams and Huckle 9-13 Multiple-Choice benchmark. This assessment contrasts the results obtained from the EQUATOR framework with those derived from non-EQUATOR evaluations. The Z Interval Error serves as a metric for quantifying the uncertainty inherent in the evaluation scores.}
\label{fig:Z Interval Error for Each Model on the Williams and Huckle 9-13 Multiple-Choice Benchmark}
\end{figure}

\begin{figure}[ht]
\centering
\includegraphics[width=1\linewidth]{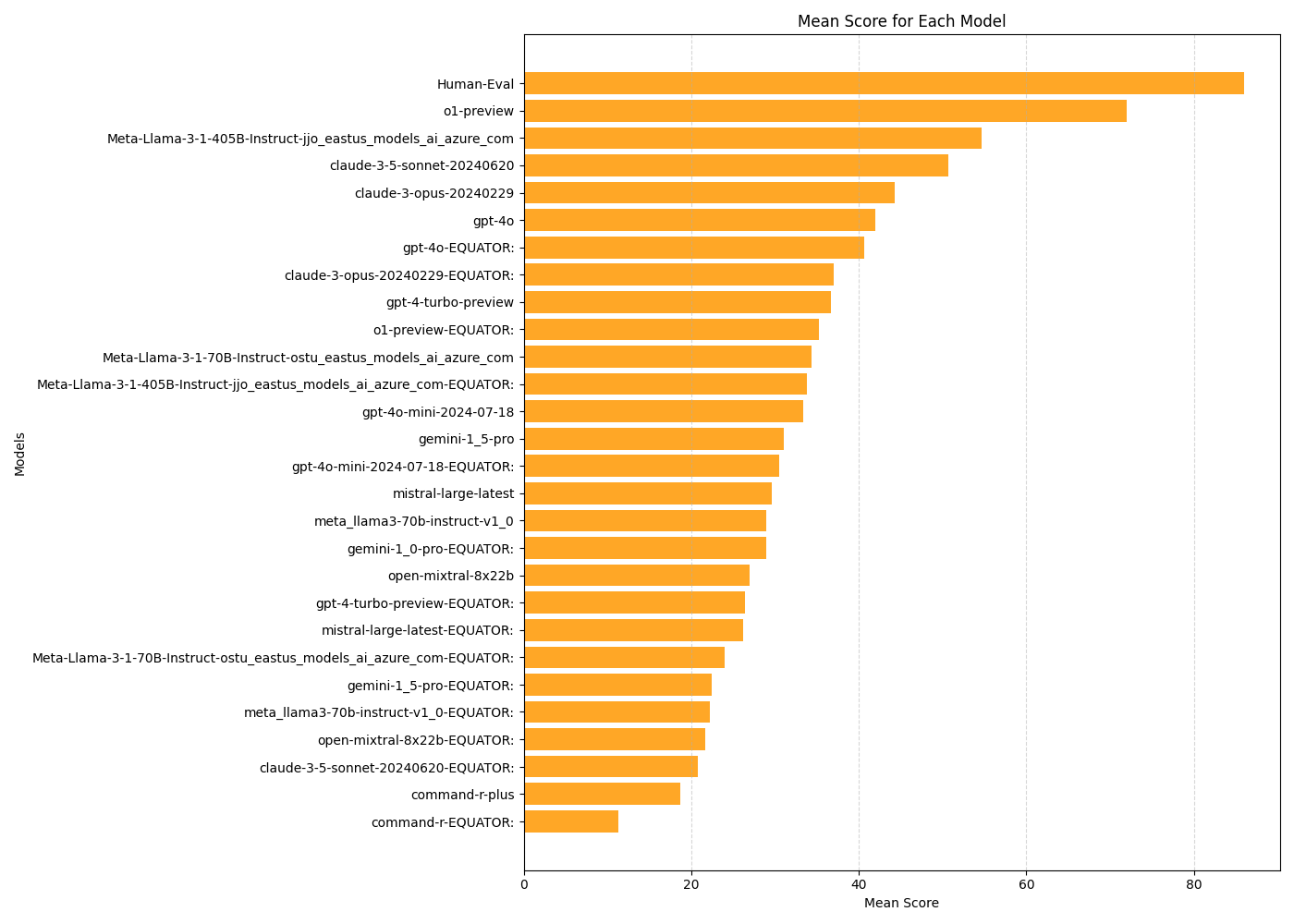}
\caption{ Mean Scores for Each Model on the Williams and Huckle 9-13 Multiple-Choice Benchmark.
This chart shows the average performance of each model, with EQUATOR scores significantly lower than the original scores, highlighting the framework's stricter evaluation standards. Human-Eval serves as a benchmark for comparison.}
\label{fig:9-13-mean}
\end{figure}

\begin{figure}[ht]
\centering
\includegraphics[width=1\linewidth]{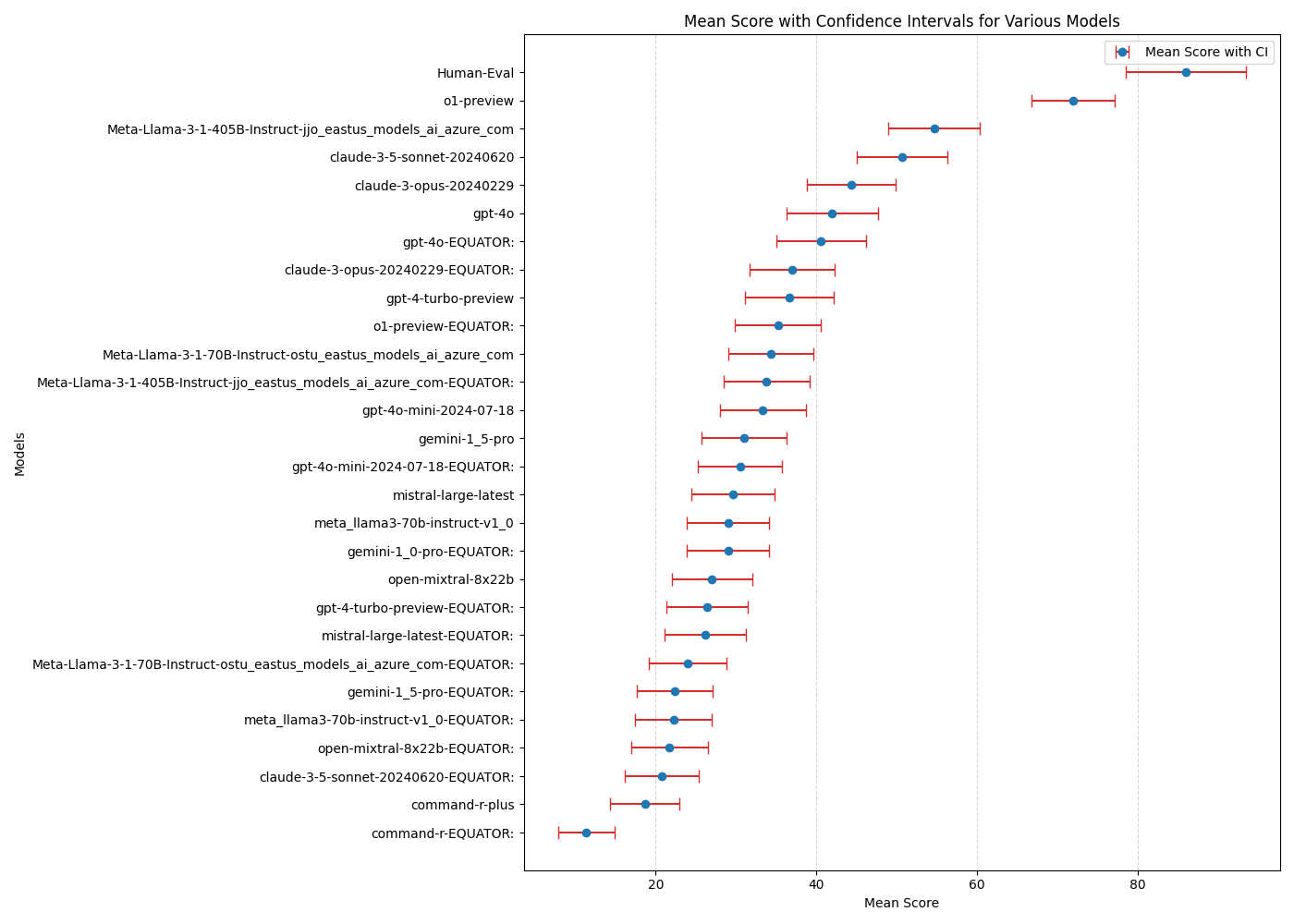}
\caption{Comparison of Mean Scores, Confidence Interval Lower and Upper Bounds for Each Model on the Williams and Huckle 9-13 Multiple-Choice Benchmark.}
\label{fig:9-13-ci}
\end{figure}

\begin{figure}[ht]
\centering
   \includegraphics[width=1\linewidth]{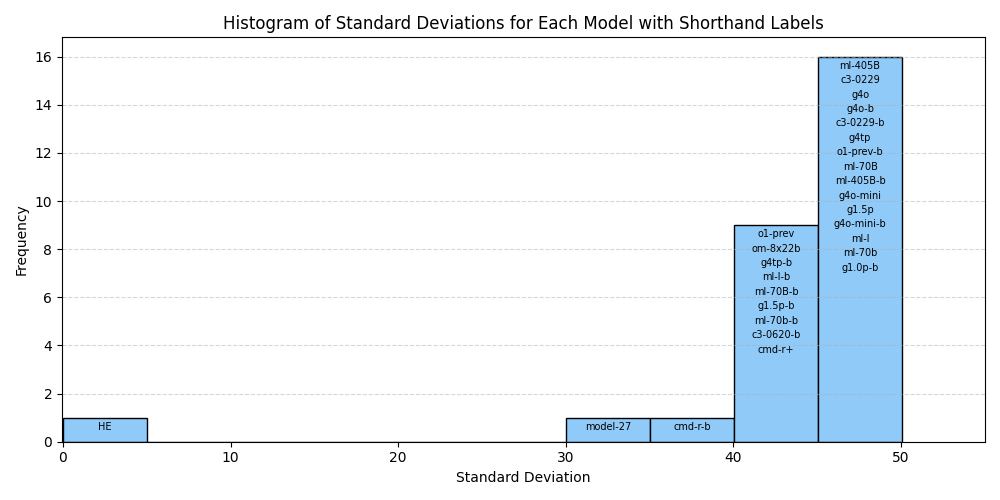}
   \includegraphics[width=1\linewidth]{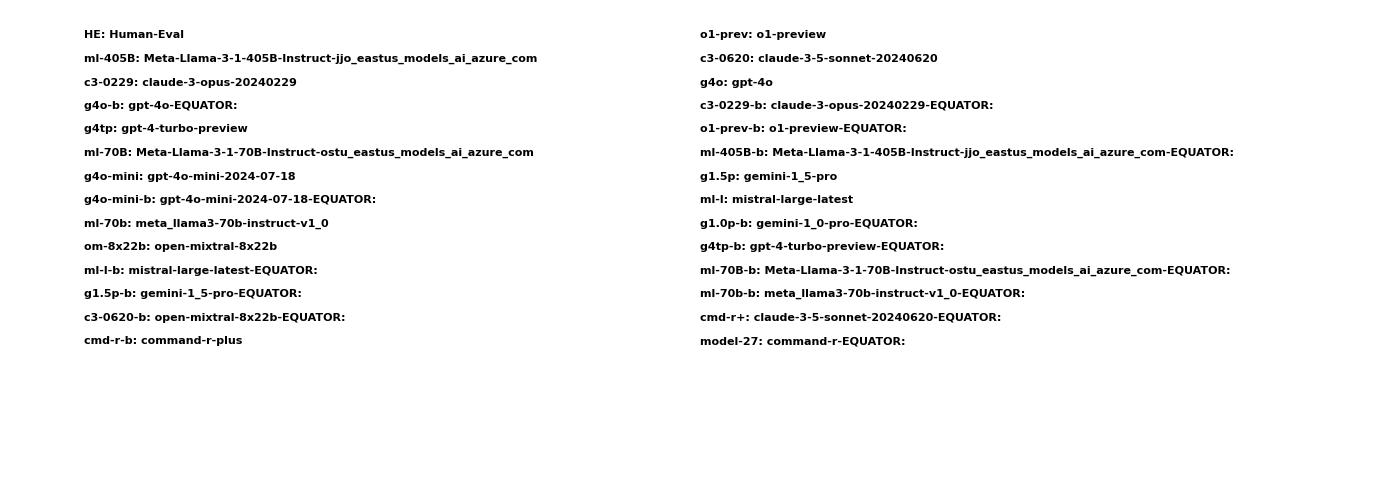}
\caption{Histogram of Standard Deviations for Each Model on the Williams and Huckle 9-13 Multiple-Choice Benchmark.
This histogram highlights the variability in model scores, with Human-Eval demonstrating minimal variability, whereas models evaluated under EQUATOR or original benchmarks show broader score distributions, indicating less consistent performance.}
\label{fig:9-13-histogram}
\end{figure}

\FloatBarrier
\newpage

\subsection{Experiment 1 - Statistical Calculations }
\label{appendix:stat9-13}

Let \( n \) be the number of paired observations (models), which is \( n = 13 \).

First, we calculate the mean of the differences (\( \bar{d} \)):

\[
\bar{d} = \frac{1}{n} \sum_{i=1}^{n} d_i
\]

Using the differences from Table~\ref{tab:mean_differences1}:

\begin{align*}
\bar{d} &= \frac{
    1.39 + 10.28 + 7.33 + 36.77 +
    10.38 + 20.88 + 2.83 + 8.59 +
    3.49 + 6.79 + 5.28 + 29.92 +
    7.33
}{13} \\
&= \frac{150.67}{13} \\
&\approx 11.59
\end{align*}

Next, we calculate the standard deviation of the differences (\( s_d \)):

\[
s_d = \sqrt{\frac{\sum_{i=1}^{n} (d_i - \bar{d})^2}{n - 1}}
\]

Compute each \( (d_i - \bar{d})^2 \):

\begin{align*}
(1.39 - 11.59)^2 &= (-10.20)^2 = 104.04 \\
(10.28 - 11.59)^2 &= (-1.31)^2 = 1.72 \\
(7.33 - 11.59)^2 &= (-4.26)^2 = 18.13 \\
(36.77 - 11.59)^2 &= (25.18)^2 = 634.09 \\
(10.38 - 11.59)^2 &= (-1.21)^2 = 1.46 \\
(20.88 - 11.59)^2 &= (9.29)^2 = 86.32 \\
(2.83 - 11.59)^2 &= (-8.76)^2 = 76.73 \\
(8.59 - 11.59)^2 &= (-3.00)^2 = 9.00 \\
(3.49 - 11.59)^2 &= (-8.10)^2 = 65.61 \\
(6.79 - 11.59)^2 &= (-4.80)^2 = 23.04 \\
(5.28 - 11.59)^2 &= (-6.31)^2 = 39.81 \\
(29.92 - 11.59)^2 &= (18.33)^2 = 336.16 \\
(7.33 - 11.59)^2 &= (-4.26)^2 = 18.13 \\
\end{align*}

The sum of squared differences:

\begin{align*}
\sum_{i=1}^{n} (d_i - \bar{d})^2 &= 104.04 + 1.72 + 18.13 + 634.09 + 1.46 + 86.32 \\
&\quad + 76.73 + 9.00 + 65.61 + 23.04 + 39.81 + 336.16 + 18.13 \\
&= 1,\!414.24
\end{align*}

Now, calculate \( s_d \):

\[
s_d = \sqrt{\frac{1,414.24}{n - 1}} = \sqrt{\frac{1,414.24}{12}} = \sqrt{117.85} \approx 10.86
\]

Calculate the standard error (\( SE \)):

\[
SE = \frac{s_d}{\sqrt{n}} = \frac{10.86}{\sqrt{13}} \approx \frac{10.86}{3.6056} \approx 3.01
\]

Compute the t-statistic:

\[
t = \frac{\bar{d}}{SE} = \frac{11.59}{3.01} \approx 3.85
\]

Degrees of freedom:

\[
df = n - 1 = 12
\]

\paragraph{Determining the p-value}

Using a t-distribution table or calculator for \( t = 3.85 \) with \( df = 12 \), we find:

\[
p \approx 0.0026
\]

\paragraph{Interpretation}

Since \( p < 0.005 \), the difference in mean scores between the traditional evaluation and the deterministic scoring framework is statistically significant. We reject the null hypothesis that there is no difference between the mean scores under the two evaluation methods.

\FloatBarrier
\newpage

\hrule 

\subsection{Experiment 2 - Graphical views scoring using the EQUATOR Evaluator on the 2024-06-12-Benchmark}

\begin{figure}[h!]
\centering
\includegraphics[width=1\linewidth]{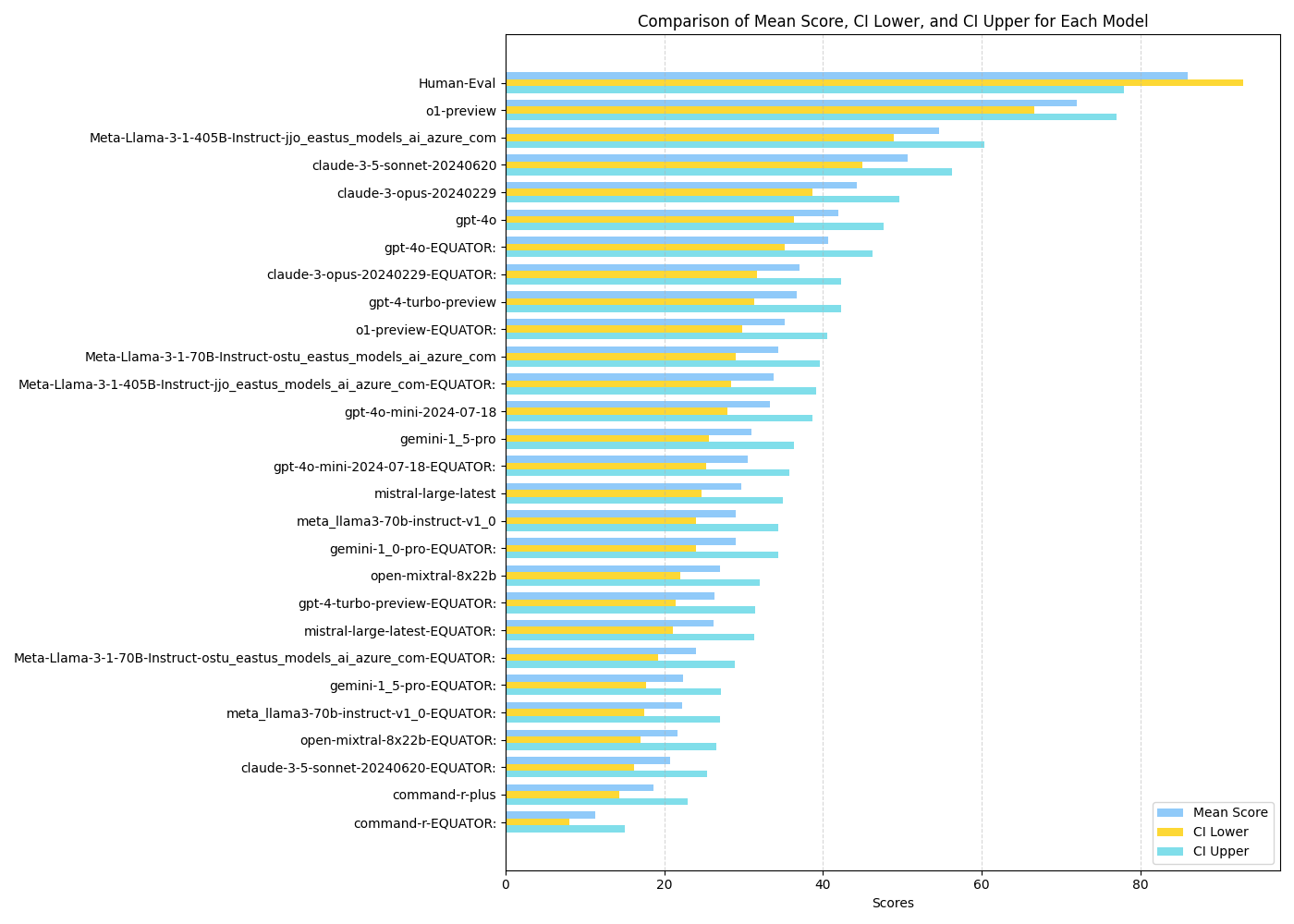}
\caption{Comparison of Mean Scores, Confidence Interval Lower and Upper Bounds for Each Model on the Williams and Huckle 6-12 Open-ended Benchmark.}
\label{fig:6-12-mci}
\end{figure}

\newpage 

\begin{figure}[h!]
\centering
\includegraphics[width=1\linewidth]{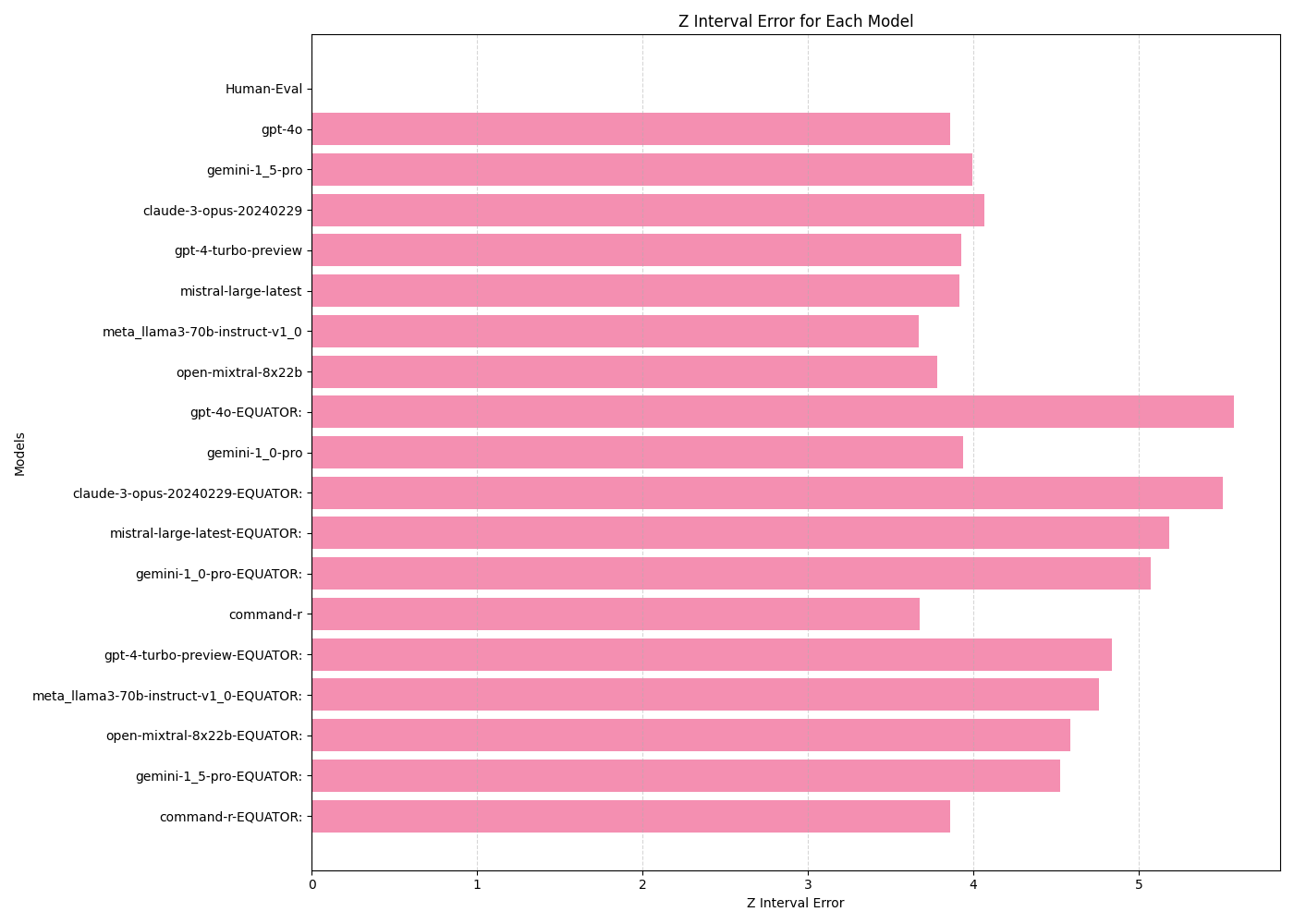}
\caption{Z Interval Error for Each Model on the Williams and Huckle 9-13 Multiple-Choice Benchmark.
This chart highlights the variability in scoring across models, with EQUATOR-scored models showing higher Z Interval Errors, indicating stricter evaluation methods and less consistency than the original benchmark scoring.}
\label{fig:6-12-z}
\end{figure}

\begin{figure}[h!]
\centering
\includegraphics[width=1\linewidth]{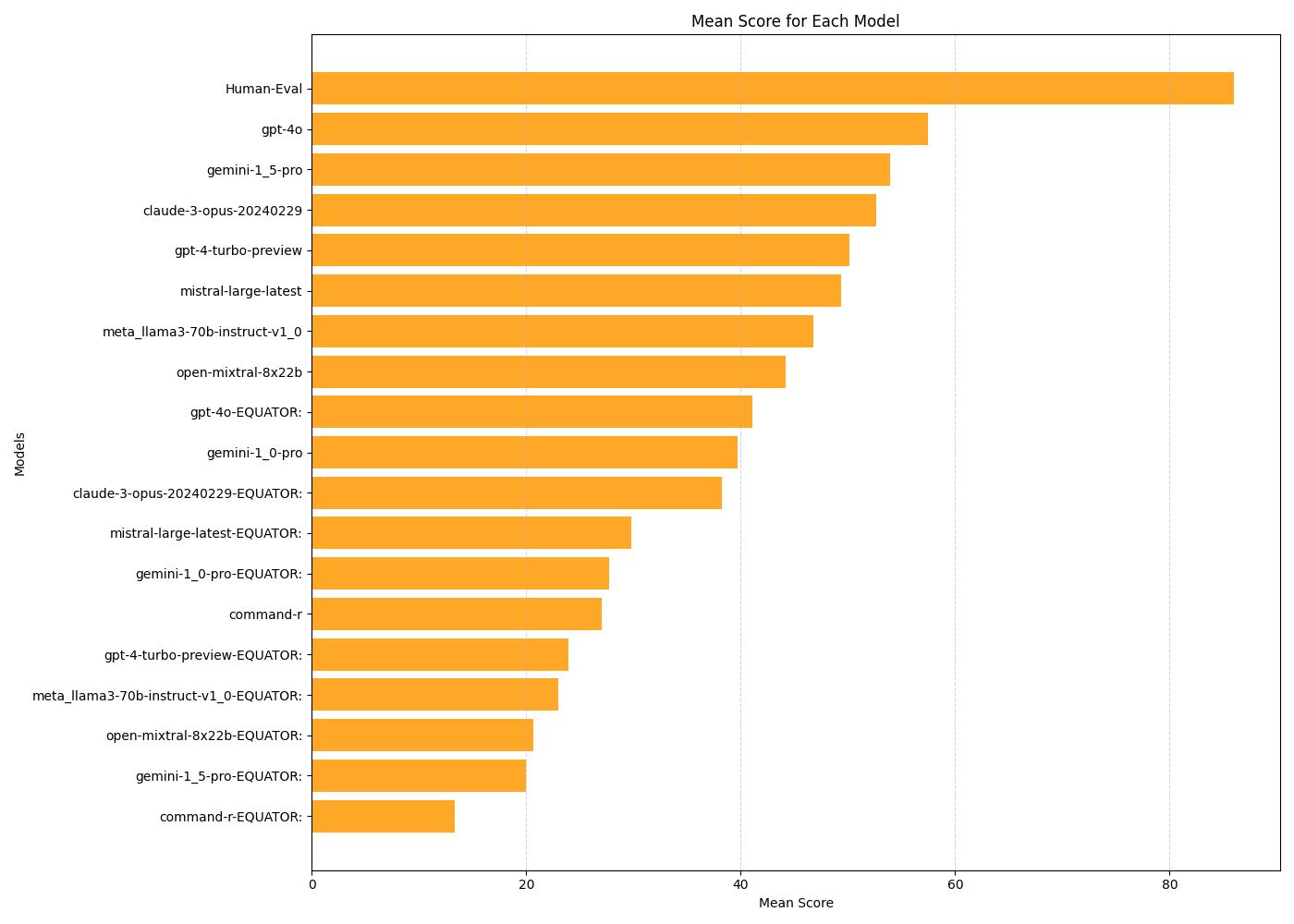}
\caption{Mean Scores for Each Model on the Williams and Huckle 9-13 Multiple-Choice Benchmark.
This figure highlights the differences in mean scores, demonstrating that EQUATOR's stricter evaluation criteria result in lower scores for models compared to the original scoring. Human-Eval remains the highest-performing benchmark.}
\label{fig:6-12-mean}
\end{figure}

\begin{figure}[h!]
\centering
\includegraphics[width=1\linewidth]{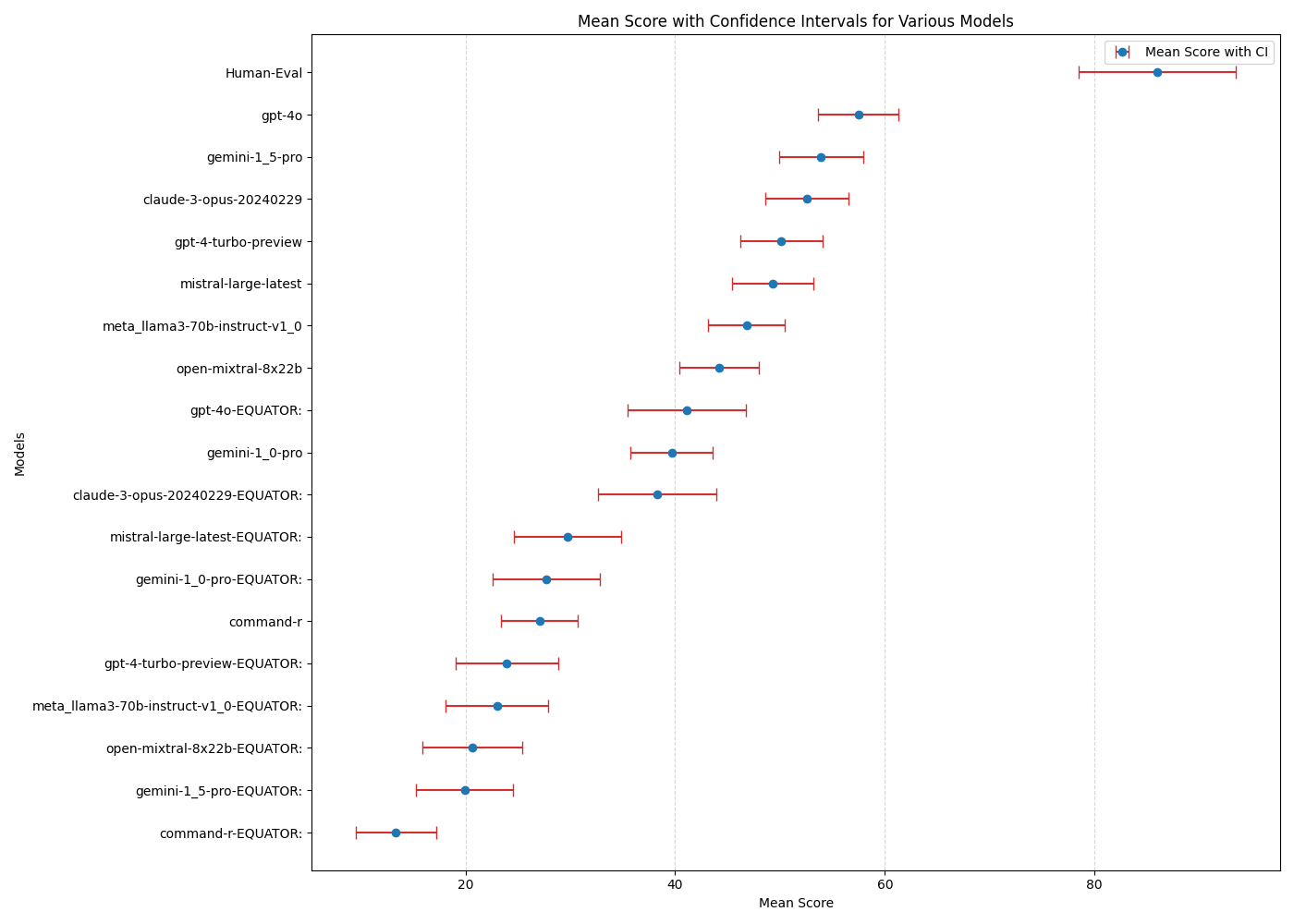}
\caption{Mean Scores with Confidence Intervals for Various Models on the Williams and Huckle 9-13 Multiple-Choice Benchmark.
This chart illustrates the mean performance of each model along with the range of variability (confidence intervals). The stricter evaluation criteria of EQUATOR result in lower scores and greater variability than the original benchmark scoring.}
\label{fig:6-12-ci}
\end{figure}

\newpage 
\begin{figure}[h!]
\centering
\includegraphics[width=.8\linewidth]{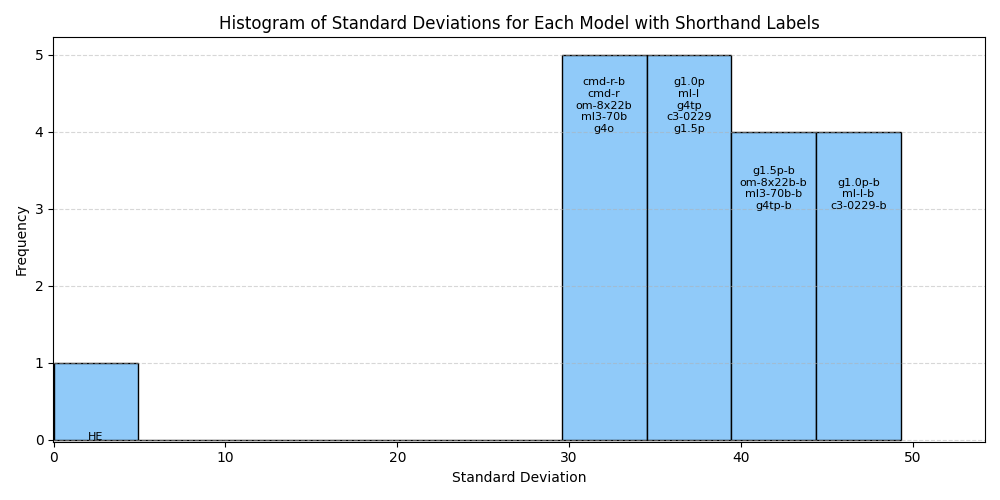}
\includegraphics[width=.8\linewidth]{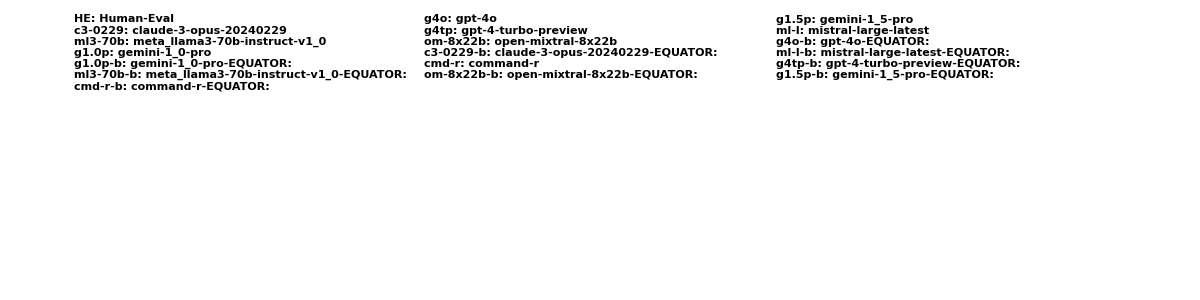}
\caption{Histogram of Standard Deviations for Each Model on the Williams and Huckle 9-13 Multiple-Choice Benchmark.
This histogram highlights the variability in performance across models, with Human-Eval showing minimal variability and other models exhibiting broader distributions, emphasizing the stricter and more variable evaluation criteria of the EQUATOR framework.  }
\label{fig:6-12-histogram}
\end{figure}

\FloatBarrier
\newpage 

\subsection{Experiment 2 -Statistical Calculations for 6-12}
\label{appendix:Stats-6-12}

Let \( n \) be the number of paired observations (models), which is \( n = 9 \).

First, we calculate the mean of the differences (\( \bar{d} \)):

\[
\bar{d} = \frac{1}{n} \sum_{i=1}^{n} d_i = \frac{183.43}{9} \approx 20.38
\]

Next, we calculate the standard deviation of the differences (\( s_d \)):

\[
s_d = \sqrt{\frac{\sum_{i=1}^{n} (d_i - \bar{d})^2}{n - 1}}
\]

Compute each \( (d_i - \bar{d})^2 \):

\begin{align*}
(16.37 - 20.38)^2 &= (-4.01)^2 = 16.09 \\
(34.00 - 20.38)^2 &= (13.62)^2 = 185.47 \\
(14.34 - 20.38)^2 &= (-6.04)^2 = 36.44 \\
(26.18 - 20.38)^2 &= (5.79)^2 = 33.57 \\
(19.57 - 20.38)^2 &= (-0.81)^2 = 0.66 \\
(23.79 - 20.38)^2 &= (3.41)^2 = 11.61 \\
(23.53 - 20.38)^2 &= (3.15)^2 = 9.93 \\
(11.96 - 20.38)^2 &= (-8.42)^2 = 70.84 \\
(13.69 - 20.38)^2 &= (-6.69)^2 = 44.82 \\
\end{align*}

Sum of squared differences:

\[
\sum_{i=1}^{n} (d_i - \bar{d})^2 = 16.09 + 185.47 + 36.44 + 33.57 + 0.66 + 11.61 + 9.93 + 70.84 + 44.82 = 409.43
\]

Calculate \( s_d \):

\[
s_d = \sqrt{\frac{409.43}{n - 1}} = \sqrt{\frac{409.43}{8}} = \sqrt{51.18} \approx 7.16
\]

Calculate the standard error (\( SE \)):

\[
SE = \frac{s_d}{\sqrt{n}} = \frac{7.16}{\sqrt{9}} = \frac{7.16}{3} \approx 2.39
\]

Compute the t-statistic:

\[
t = \frac{\bar{d}}{SE} = \frac{20.38}{2.39} \approx 8.55
\]

Degrees of freedom:

\[
df = n - 1 = 8
\]

\paragraph{Determining the p-value}

Using a t-distribution table or calculator for \( t = 8.55 \) with \( df = 8 \), we find:

\[
p < 0.0001
\]

\paragraph{Interpretation}

Since \( p < 0.0001 \), the difference in mean scores between the traditional evaluation and the deterministic scoring framework is statistically significant. We reject the null hypothesis that there is no difference between the mean scores under the two evaluation methods.

\FloatBarrier
\newpage

\subsection{Cohen d Background}
\subsubsection{1. \textbf{Quantifying the Impact of Scoring Methods}}

\cite{cohen1988statistical}
\subsubsection{\textbf{1. Cohen, Jacob (1988). Statistical Power Analysis for the Behavioral Sciences (2nd Edition).}}

This seminal book is the cornerstone of statistical power analysis, offering a comprehensive guide to understanding and applying statistical methods in behavioral research. Cohen introduces the concept of \textbf{effect size} as a practical measure of the magnitude of differences between groups, which is independent of sample size.

\paragraph{Key Contributions:}

\begin{itemize}
    \item \textbf{Cohen's d}: A specific measure of effect size used to quantify the difference between two group means in terms of standard deviation units. Cohen also proposed benchmarks for interpreting effect size: 
    \begin{itemize}
        \item \textbf{Small} (d = 0.2): Minor difference.
        \item \textbf{Medium} (d = 0.5): Noticeable but not large.
        \item \textbf{Large} (d = 0.8): Substantial difference.
    \end{itemize}
    \item Emphasizes that statistical significance alone is insufficient for meaningful research, advocating for the combined use of effect size and power analysis.
    \item Provides work examples and recommendations for designing experiments with adequate statistical power.
\end{itemize}

According to Cohen's conventions:

\begin{itemize}
    \item Small effect: \( d = 0.2 \)
    \item Medium effect: \( d = 0.5 \)
    \item Large effect: \( d = 0.8 \)
\end{itemize}

\cite{cohen1969statistical}
\subsubsection{\textbf{2. Cohen, Jacob (1969). Statistical Power Analysis for the Behavioral Sciences (1st Edition).}}

This earlier edition laid the foundation for statistical power analysis as a critical part of experimental design and evaluation. Cohen introduced statistical power as the probability of detecting a true effect in a study and emphasized the importance of considering both Type I (false positives) and Type II (false negatives) errors.

\paragraph{Key Contributions:}

\begin{itemize}
    \item Introduced the idea of \textbf{power analysis} to ensure studies are designed with an adequate sample size to detect meaningful effects.
    \item First \textbf{, Cohen's d} was introduced as a practical way to measure effect sizes for t-tests, standardizing the magnitude of differences for comparison across studies.
    \item Demonstrated the practical implications of effect size on the interpretation of results, advocating for its use alongside p-values.
\end{itemize}

\cite{sawilowsky2009new}
\subsubsection{\textbf{3. Sawilowsky, Shlomo S. (2009). New Effect Size Rules of Thumb.}}

This article revisited Cohen's original benchmarks for effect sizes and proposed updated guidelines based on more recent empirical studies and expanded use cases.

\paragraph{Key Contributions:}

\begin{itemize}
    \item Highlights the limitations of Cohen's benchmarks in certain contexts, such as education, medicine, and other applied sciences.
    \item Proposes \textbf{new rules of thumb} for interpreting effect sizes that are more tailored to specific research fields.
    \item Encourages researchers to consider context and domain-specific norms when interpreting effect sizes, rather than rigidly adhering to generic benchmarks.
\end{itemize}

\cite{lakens2013calculating}
\subsubsection{\textbf{4. Lakens, Daniël (2013). Calculating and Reporting Effect Sizes to Facilitate Cumulative Science.}}

This article provides a practical guide for researchers to compute, interpret, and report effect sizes, such as Cohen’s d, in statistical analyses. Lakens emphasizes the importance of effect sizes for replicable and cumulative scientific progress.

\paragraph{Key Contributions:}

\begin{itemize}
    \item Details the calculation of Cohen's d for different statistical tests (e.g., t-tests, ANOVA).
    \item Encourages transparent reporting of effect sizes and confidence intervals to improve the interpretability and reproducibility of findings.
    \item Advocates for using effect sizes as a complement to p-values, making results more meaningful and practical for cumulative research.
\end{itemize}

\subsubsection{\textbf{Summary of Contributions to Our Use Case}}

These references provide a theoretical and practical foundation for using Cohen's d in our evaluation of LLMs:

\begin{enumerate}
    \item \textbf{Cohen’s Work}: Established the importance of effect size as a measure of practical significance and created interpretable benchmarks that can help quantify the impact of scoring methods in our experiments.
    \item \textbf{Sawilowsky’s Updates}: Suggest that domain-specific considerations may influence the interpretation of effect sizes, reinforcing the importance of contextual analysis.
    \item \textbf{Lakens’ Practical Guide}: Offers detailed methodologies for calculating and reporting effect sizes, ensuring rigor and reproducibility in our work.
\end{enumerate}
By leveraging these works, our approach to using Cohen’s d to evaluate deterministic scoring frameworks becomes both statistically robust and practically meaningful. Let me know if you'd like this adapted further!

\subsubsection{7. \textbf{Communicating Results to Stakeholders}}

\textbf{Clarity for Non-Experts}: Cohen’s d is widely understood and provides an intuitive way to communicate evaluation outcomes. Stakeholders can easily grasp terms like: 

\begin{itemize}
    \item \textbf{Small Effect (d = 0.2)}: Minor differences between scoring methods.
    \item \textbf{Medium Effect (d = 0.5)}: Moderate impact of deterministic scoring.
    \item \textbf{Large Effect (d = 0.8 or more)}: Significant improvement or shift in evaluation.
\end{itemize}

This aids in:

\begin{itemize}
    \item Demonstrating the importance of adopting the deterministic framework.
    \item Justifying resource allocation for further improvements based on clear, interpretable evidence.
\end{itemize}

\newpage

\subsection{Cohen's D Chart Analysis}\label{appendix:Chart Analysis}

The horizontal bar chart visualizes the \textbf{Cohen’s d effect sizes} for each model compared to the EQUATOR evaluator (referred to as "Human-Eval" in the dataset). It quantifies how much the performance of each model differs from the EQUATOR evaluator in terms of standard deviation units. The chart uses a custom color scheme to enhance readability and highlight differences across models.

\subsubsection{\textbf{Key Components of the Chart}}

\paragraph{\textbf{1. Horizontal Bars}}

\textbf{Bars Represent Cohen’s d}: Each bar represents Cohen's d value for a model. 

\begin{itemize}
    \item \textbf{Length of the Bar}: The longer the bar, the greater the effect size, meaning the model's performance deviates more significantly from the EQUATOR evaluator.
    \item \textbf{Directionality}: Positive Cohen’s d values (all in this case) indicate that the EQUATOR evaluator outperformed the corresponding model.
\end{itemize}

\paragraph{\textbf{3. Axes}}

\begin{itemize}
    \item \textbf{Horizontal Axis (Cohen's d)}:
    \begin{itemize}
        \item Represents the magnitude of the difference in performance between the EQUATOR evaluator and each model.
        \item Higher values mean a larger difference between the mean scores, standardized by the pooled variability.
    \end{itemize}
    \item \textbf{Vertical Axis (Models)}: 
    \begin{itemize}
        \item Lists the models compared against the EQUATOR evaluator.
        \item Models are sorted by ascending Cohen’s d values for clarity and readability.
    \end{itemize}
\end{itemize}

\subsection*{Factors to Consider:}
\label{appendix:stat-sig}
\begin{enumerate}
  \item \textbf{Desired Confidence Level}: Typically, a 95\% confidence level is used in educational assessments.
  \item \textbf{Margin of Error}: This is the acceptable error range (e.g., ±5\% or ±10\%).
  \item \textbf{Variability}: The level of variance in responses; higher variability means more questions are needed to capture a reliable measure.
  \item \textbf{Population Size}: For smaller sample groups, you need more questions to achieve significance.
  \item \textbf{Effect Size}: How large of a difference or effect you expect to detect. Larger effects require fewer questions.
\end{enumerate}

\subsubsection{\textbf{Interpreting the Chart}}

\paragraph{\textbf{The magnitude of Cohen’s d}}

Cohen’s d values can be interpreted based on standard guidelines:

\begin{itemize}
    \item \textbf{Small Effect (d = 0.2)}: Minimal difference between EQUATOR and the model.
    \item \textbf{Medium Effect (d = 0.5)}: Moderate difference, potentially meaningful.
    \item \textbf{Large Effect (d = 0.8 or higher)}: Significant difference, indicating substantial deviation from the EQUATOR Evaluator's performance.
\end{itemize}

\paragraph{\textbf{Insights from the Chart}}

\begin{enumerate}
    \item \textbf{Top Models (Lowest Cohen’s d)}:
    \begin{itemize}
        \item Models near the bottom of the chart (shorter bars) have smaller effect sizes, meaning their performance is closer to the EQUATOR evaluator.
        \item These models are relatively aligned with the EQUATOR evaluator, suggesting they produce similar results.
    \end{itemize}
    \item \textbf{Outperforming Models (Higher Cohen’s d)}:
    \begin{itemize}
        \item Models near the top of the chart (longer bars) show the largest deviations from the EQUATOR evaluator.
        \item These models perform significantly worse under the given evaluation, as indicated by large positive Cohen’s d values.
    \end{itemize}
    \item \textbf{Performance Variation}:
    \begin{itemize}
        \item The length of the bars demonstrates the variability in how well the models align with the EQUATOR evaluator.
        \item A wide range of Cohen’s d values (e.g., from \~0.3 to >2.0) indicates substantial variability in model performance.
    \end{itemize}
\end{enumerate}

\paragraph{\textbf{Example Observations}:}

\begin{itemize}
    \item \textbf{Lowest Effect Size}: The first bar represents a model with the smallest deviation from EQUATOR, suggesting its performance is relatively close.
    \item \textbf{Highest Effect Size}: The final bar displays the model with the highest deviation, indicating it performs significantly worse than the EQUATOR evaluators.

\end{itemize}

\subsubsection{\textbf{Why This Chart Matters}}

\begin{enumerate}
    \item \textbf{Objective Comparison}:
Cohen’s d standardizes the difference in mean scores, allowing for an unbiased comparison across models, irrespective of score scale or variability.

    \item \textbf{Highlights Model Robustness}:
Models with lower Cohen’s d values are more consistent with the EQUATOR evaluator, suggesting better alignment with desired evaluation criteria.

    \item \textbf{Prioritization}:
The chart helps identify models that need improvement (those with high Cohen’s d values) and those performing well (low Cohen’s d).

\end{enumerate}

\newpage

\section{Formula to Estimate Sample Size (Number of Questions):}
You can use a \textbf{sample size formula} commonly used in statistics:
\[
n = \frac{Z^2 \cdot p \cdot (1-p)}{E^2}
\]
Where:
\begin{itemize}
  \item \( n \) = required sample size (number of questions)
  \item \( Z \) = Z-score, based on the desired confidence level (1.96 for 95\% confidence)
  \item \( p \) = estimated proportion of correct responses (0.5 is a safe assumption if unsure)
  \item \( E \) = margin of error (e.g., 0.05 for ±5\%)
\end{itemize}

\subsection*{Example:}
Confidence Level: 95\% (\( Z = 1.96 \)) Estimated proportion of correct responses: 50\% (\( p = 0.5 \))  
Margin of Error: ±5\% (\( E = 0.05 \))

\[
n = \frac{(1.96)^2 \cdot 0.5 \cdot (1-0.5)}{(0.05)^2} = \frac{3.8416 \cdot 0.25}{0.0025} = 384.16
\]

So, approximately \textbf{385 questions} are needed to achieve a statistically significant result with a 95\% confidence level and a 5\% margin of error.

\end{document}